\documentclass[11pt,draftcls,onecolumn]{IEEEtran}

\usepackage{cite}

\usepackage{colortbl}
\usepackage{color}
\usepackage{bm}
\usepackage{bbm}
\usepackage{amsmath}
\usepackage{multirow}
\usepackage{subcaption}

\usepackage{cases}
\usepackage{amssymb}
\usepackage{amsthm}
\usepackage{graphicx}
\usepackage{mathrsfs}
\usepackage{empheq}	
\usepackage[mathcal]{euscript}
\usepackage[margin = 2cm]{geometry}
\usepackage{framed}
\usepackage{comment}

\usepackage{algorithm}
\usepackage{algorithmic}

\usepackage{dsfont}
\usepackage{xr}
\makeatletter
\newcommand*{\addFileDependency}[1]{% argument=file name and extension
  \typeout{(#1)}
  \@addtofilelist{#1}
  \IfFileExists{#1}{}{\typeout{No file #1.}}
}
\makeatother

\DeclareFontFamily{U}{mathx}{\hyphenchar\font45}
\DeclareFontShape{U}{mathx}{m}{n}{
      <5> <6> <7> <8> <9> <10>
      <10.95> <12> <14.4> <17.28> <20.74> <24.88>
      mathx10
      }{}
\DeclareSymbolFont{mathx}{U}{mathx}{m}{n}
\DeclareFontSubstitution{U}{mathx}{m}{n}
\DeclareMathAccent{\widecheck}{0}{mathx}{"71}
\newtheorem{theorem}{Theorem}

\newtheorem{corollary}{Corollary}

\newtheorem{assumption}{Assumption}

\newtheorem{property}{Property}

\renewcommand{\P}{\mathbb{P}}
\newcommand{\E}{\mathbb{E}}

\newcommand{\beq}{\begin{equation}}
\newcommand{\eeq}{\end{equation}}
\newcommand{\beqa}{\begin{IEEEeqnarray}{rCl}}
\newcommand{\eeqa}{\end{IEEEeqnarray}}

\DeclareMathOperator*{\argmax}{arg\,max}
\DeclareMathOperator*{\argmin}{arg\,min}

\DeclareMathOperator*{\asl}{\textnormal{A}\!\!\!\!\phantom{a}^2\textnormal{SL}}

\title{Doubly Adaptive Social Learning}
\author{Marco Carpentiero, Virginia Bordignon, Vincenzo Matta, and Ali H. Sayed
\thanks{
Marco Carpentiero and Vincenzo Matta are with the Department of Information and Electrical Engineering and Applied Mathematics (DIEM), University of Salerno, via Giovanni Paolo II, I-84084, Fisciano (SA), Italy, and Vincenzo Matta is also with the National Inter-University Consortium for Telecommunications (CNIT), Italy (e-mails: \{mcarpentiero, vmatta\}@unisa.it).

The work of V. Matta was partially supported by the European Union under the Italian National Recovery and Resilience Plan (NRRP) of NextGenerationEU, partnership on ``Telecommunications of the Future'' (PE00000001 - program ``RESTART'').

This work was produced while Virginia Bordignon was a post-doc with the \'Ecole Polytechnique F\'ed\'erale de Lausanne EPFL, School of Engineering, CH-1015 Lausanne, Switzerland (e-mail: virginia.bordignon@alumni.epfl.ch).

Ali H. Sayed is with the \'Ecole Polytechnique F\'ed\'erale de Lausanne EPFL, School of Engineering, CH-1015 Lausanne, Switzerland (e-mail: ali.sayed@epfl.ch).

A short conference version of this article was presented at ICASSP 2024~\cite{A2SLICASSP2024}.
}
}

\begin{document}
	\maketitle

\begin{abstract}
In social learning, a network of agents assigns probability scores
(beliefs) to some hypotheses of interest, which rule the generation of local streaming data observed by each agent. Belief formation takes place by means of an iterative two-step procedure where: $i)$ the agents update locally their beliefs by using some likelihood model; and $ii)$ the updated beliefs are combined with the beliefs of the neighboring agents, using a pooling rule. This procedure can fail to perform well in the presence of dynamic drifts, leading the agents to incorrect decision making. Here, we focus on the {\em fully online} setting where both the true hypothesis and the likelihood models can change over time. We propose the doubly adaptive social learning ($\asl$) strategy, which infuses social learning with the necessary adaptation capabilities. This goal is achieved by exploiting two adaptation stages: $i)$ a stochastic gradient descent update to learn and track the drifts in the decision model; $ii)$ and an adaptive belief update to track the true hypothesis changing over time. These stages are controlled by two adaptation parameters that govern
the evolution of the error probability for each agent. We show that all agents learn consistently for sufficiently small adaptation parameters, in the sense that they ultimately place all their belief mass on the true hypothesis. In particular, the probability of choosing the wrong hypothesis converges to values on the order of the adaptation parameters. The theoretical analysis is illustrated both on synthetic data and by applying the $\asl$ strategy to a social learning problem in the online setting using real data. 
\end{abstract}


\section{Introduction and Related Work}
Social learning embodies a family of popular algorithms employed for collaborative opinion formation, where a group of agents assigns probability scores ({\em beliefs}) to some hypotheses of interest, based on the observation of private streaming data and the beliefs exchanged with their neighbors~\cite{SLMagazine2024, SLBook, ChamleyBook, acemoglu2011opinion, acemoglu2011bayesian, chamley2013models, krishnamurthy2013social, mossel2017opinion, zhao2012learning}. 
In their traditional implementations, social learning algorithms have been thoroughly examined in the literature, and have been shown to offer provable learning guarantees. Under reasonable technical conditions, each agent ends up placing all the probability mass on the true underlying hypothesis that gives rise to the data~\cite{SLMagazine2024, SLBook, zhao2012learning, jadbabaie2012non, ShahrampourTAC2016, nedic2017, molavi2018theory, lalitha2018}.

These learning guarantees are provided for a {\em static} learning setting. In contrast, an {\em online} setting requires the social learning strategy to be able to react and {\em adapt} promptly to the two inherent sources of non-stationarity of the inferential problem: $(i)$ {\em drifts in the true hypothesis}, e.g., the true state of nature is shifting and the social learning paradigm must track its variations; $(ii)$ {\em drifts in the likelihood models} underlying the social learning algorithm, e.g., the likelihood ratios that each agent adopts to extract information from the data must be updated in face of new evidence accumulated in the observed data. Poor adaptation capabilities in any of the two aforementioned domains can lead to poor performance and inconsistency, since the agents are simply not able to correctly detect the true state of nature by lacking perception of the dynamics underlying the inferential problem.  

Traditional social learning does not exhibit any adaptation either to hypotheses or model drifts, and is therefore unreliable in online settings. Regarding drifts in the hypotheses, the exponential convergence of the beliefs~\cite{nedic2017, molavi2018theory, lalitha2018} makes the agents {\em stubborn}: once deciding for the true hypothesis with high confidence, in the presence of a change in that hypothesis, they get stuck into their past determination and undergo long delays in responding to the change. Regarding drifts in the models, traditional social learning assumes that the models are known beforehand and no adaptation mechanism is in place to track their variations. 

The issue of drifting hypotheses can be managed by the recently proposed {\em adaptive} social learning (ASL) paradigm~\cite{bordignon2021adaptive}, which infuses traditional social learning with adaptation by means of an {\em adaptive update step}. In the ASL strategy, the agents perform updates of their beliefs that take into account the relative importance assigned to past and new data, leading to the desired tracking capabilities regarding the true hypothesis. Reliable learning is attained even in the presence of hypotheses drifts, albeit under prior knowledge of the likelihood models. This prevents the application of the ASL strategy in fully online scenarios, where the models may be unknown and varying over time. 

The issue of unknown models has been recently addressed in the context of social {\em machine} learning (SML)~\cite{BordignonVlaski2023,HuBordignon2023} and social learning with uncertain models~\cite{HareUribeTSP2020}. 
In these works, the agents circumvent the lack of the exact likelihood models by taking advantage of {\em training data}, used to learn the decision models (e.g., the likelihood ratios), which are then employed to perform social learning over a stream of {\em prediction data}.
However, while these approaches remove the demanding need for prior knowledge of likelihood models, they still do not provide adaptation to drifs in the model, since training is performed {\em offline}, without taking into account the dynamics of the learning problem over time.

Our goal is to infuse social learning with adaptation capabilities in both non-stationarity domains. To this aim, we propose the {\em doubly adaptive} social learning strategy, nicknamed $\asl$, which is designed to enable adaptation with respect to {\em hypotheses and model drifts}, and is therefore targeted to {\em fully online} applications. Once turned on, the $\asl$ strategy can run virtually forever, with no need of resets or re-tuning stages, since it automatically adapts to variations in the training or prediction data. Differently from the simplifying assumption often employed in learning applications, we address the realistic setting where the training and prediction phases are not separated. In our model, these phases stay concurrently active, and at each time instant new training or prediction data can be observed. Furthermore, we allow these two processes to be {\em asynchronous}, meaning that at a given time instant we can have one fresh training sample and/or one fresh prediction sample, or no samples at all. This additional flexibiliy turns out to be important to model real-world applications.

The $\asl$ strategy adds adaptation capabilities to social learning by using two adaptation stages: the first stage is a stochastic gradient descent (SGD) algorithm, which features a {\em constant} step-size as a {\em training adaptation parameter} to perform online model training. The second stage is an adaptive belief update, ruled by a {\em prediction adaptation parameter} to balance old knowledge (stored in the past beliefs) and new knowledge (extracted from the current prediction data).
We will show that the $\asl$ strategy achieves consistent learning, in the sense that, at the steady-state and for sufficiently small adaptation parameters, the probability that (at any agent) the maximum belief mass concentrates on the correct hypothesis stays close to $1$. In particular, we establish that the probability of error of each agent exhibits a vanishing transient phase with an exponential rate depending on the adaptation parameters, and a steady-state phase with an error on the order of the adaptation parameters. In accordance with adaptation theory, this result shows the trade-off between learning accuracy and learning speed, which can be exploited to address the demanding fully online setting.
We validate the theoretical analysis first on synthetic data, and then by employing the $\asl$ strategy to solve a distributed classification problem in a fully online setting with real data from the CIFAR-10 data set\cite{cifar10}.

{\bf Notation}. We denote random variables with bold font. All vectors are column vectors.
The operator $\textnormal{col}\{\cdot\}$ stacks its column-vector entries into a single column vector. 
For a nonnegative function $g(y)$ with positive argument $y$, the notation $g(y)=O(y)$ means that $g(y)\leq c\,y$ for all $y\leq y_0$, for some positive values $c$ and $y_0$.
The symbols $\E$ and $\P$ denote expectation and probability, respectively.
The superscript $^\star$ denotes quantities relative to the \emph{true} distribution of the data; the superscript $^o$ denotes quantities resulting as the solution of some \emph{optimization} process; and the superscript $'$ denotes random variables contained in the \emph{training set}, to distinguish them from the random variables generated during the prediction phase.

\section{Background}
Consider a set of $H$ hypotheses $\Theta = \{\theta_1, \theta_2, \ldots, \theta_{H}\}$.
At each time $t=1,2,\ldots$, each agent $k =1,2,\ldots,K$ observes some {\em prediction data} $\bm{x}_{k,t} \in \mathbb{R}^{M_k}$, and 
has access to likelihood models that relate the hypotheses to the data:
\beq
\ell^{\star}_{k}(x_k|\theta),\qquad \textnormal{for $x_k\in\mathbb{R}^{M_k}$ and $\theta\in\Theta$}.
\eeq
Strictly speaking, the function $\ell^{\star}_{k}(x_k|\theta)$ is a likelihood when regarded as a function of $\theta$ for a fixed $x_k$. 
For a fixed $\theta$, instead, it represents the generative model of the data $x_k$ corresponding to that $\theta$, e.g., it can be a probability mass function (pmf) or a probability density function (pdf).

In order to form their opinion on the state of nature represented by the hypotheses in the set $\Theta$, the agents assign a probability score to each $\theta \in \Theta$. The probability scores form the {\em belief vector} of agent $k$ at time $t$, which is defined as $\bm{\mu}_{k,t} = [\bm{\mu}_{k,t}(\theta_1),\bm{\mu}_{k,t}(\theta_2),\ldots,\bm{\mu}_{k,t}(\theta_H)]$, with
\begin{equation}
\bm{\mu}_{k,t}(\theta)\geq 0,\qquad \sum_{\theta\in\Theta} \bm{\mu}_{k,t}(\theta)=1.
\end{equation}
According to the traditional social learning paradigm, the belief vectors are iteratively formed applying the following two-step strategy, after proper initialization of the beliefs with some deterministic vectors $\mu_{k,0}$~\cite{SLBook, nedic2017, lalitha2018, molavi2018theory}:
\begin{subequations}
\begin{align}
\bm{\psi}_{k,t}(\theta) & = 
\frac{\bm{\mu}_{k,t-1}(\theta) \ell^{\star}_{k}(\bm{x}_{k,t}|\theta)}
{\sum\limits_{\tau \in \Theta}\bm{\mu}_{k,t-1}(\tau)\ell^{\star}_{k}(\bm{x}_{k,t}|\tau)} \nonumber \\
& \propto  \bm{\mu}_{k,t-1}(\theta) 
e^{d^{\star}_{k,t}(\bm{x}_{k,t};\theta)} ,
\label{eq:intSLStep0}
\\
\bm{\mu}_{k,t}(\theta) & \propto \prod_{j=1}^K [\bm{\psi}_{j,t}(\theta)]^{a_{jk}},
\label{eq:socialLearningStep0}
\end{align}
\end{subequations}
where the symbol $\propto$ hides the proportionality constant necessary to make $\bm{\psi}_{k,t}=[\bm{\psi}_{k,t}(\theta_1),\bm{\psi}_{k,t}(\theta_2),\ldots,\bm{\psi}_{k,t}(\theta_H)]$ and $\bm{\mu}_{k,t}$ probability vectors. The RHS of \eqref{eq:intSLStep0} is obtained by dividing $\ell^{\star}_k(\bm{x}_{k,t}|\theta)$ and $\ell^{\star}_k(\bm{x}_{k,t}|\tau)$ by $\ell^{\star}_{k}(\bm{x}_{k,t}|\theta_H)$ and introducing the {\em decision statistic or model}:\footnote{For nonsingular problems, the likelihood ratio is almost-surely nonzero.}
\beq 
d^{\star}_{k}(x_k;\theta) \triangleq \log \frac{\ell^{\star}_{k}(x_k|\theta)}{\ell^{\star}_{k}(x_k|\theta_H)}.
\label{eq:logLikDef}
\eeq
Without loss of generality, we divide by $\ell^{\star}_{k}(x_k|\theta_H)$, but we can use any other hypothesis as a ``pivot". Note that $d^{\star}_k(x_k;\theta_H)=0$ by definition. 

Step \eqref{eq:intSLStep0} produces an {\em intermediate belief} $\bm{\psi}_{k,t}(\theta)$ by performing a {\em Bayesian update} based on the new local observation $\bm{x}_{k,t}$.  In step \eqref{eq:socialLearningStep0}, each agent implements a pooling rule to combine the intermediate beliefs of the other agents. 
Specifically, agent $k$ computes a weighted geometric average where the belief of agent $j$ is raised to a weight $a_{jk}$. 

The weights $a_{jk}$ are related to the network structure that connects the agents. This network can be represented as a weighted directed graph, whose edge weights are collected into the nonnegative {\em combination matrix} $A = [a_{jk}] \in \mathbb{R}^{K \times K}$. 
The matrix $A$ must be left-stochastic~\cite{Sayed,sayednewbooks}:
\beq
a_{\ell k} \geq 0, \qquad \sum_{j=1}^K a_{\ell k} = 1.
\label{combWeights}
\eeq
When $a_{jk}=0$, agent $k$ does not receive information from agent $j$. Conversely, agent $k$ aggregates the beliefs received from the agents $j$ for which $a_{jk}>0$, which are called {\em neighbors} of $k$. 
Over the weighted graph, a {\em path} of length $n$ originating at agent $j$ and terminating at agent $k$ is identified by a sequence of nonzero weights 
\beq 
a_{j h_1}, a_{h_1 h_2}, \ldots,a_{h_n k}.
\label{eq:path}
\eeq 
We assume the following standard conditions on the paths connecting any two agents, i.e., on the combination matrix:

\begin{assumption}[{\bf Irreducible combination matrix~\cite{bib:matrix, Meyer}}]
\label{ass:irreducibleMatrix}
The combination matrix $A$ is irreducible when for any two distinct agents $j$ and $k$, there exist paths linking them in both directions. From the Perron-Frobenius theorem, an irreducible combination matrix $A$ has a single eigenvalue $\lambda$ equal to its spectral radius. With proper scaling, the eigenvector of $A$ corresponding to $\lambda$, called the Perron vector and denoted by $v = [v_1, v_2, \ldots, v_K]^{\top}$, can be normalized such that 
\beq 
Av = v, \quad \mathds{1}^{\top}v = 1, \quad v_k > 0, \;\; \textnormal{for} \;\; k = 1,2,\ldots,K.
\label{eq:perronProp}
\eeq 
~\hfill$\square$ 
\end{assumption}

\begin{assumption}[{\bf Primitive combination matrix~\cite{bib:matrix, Meyer}}] 
\label{ass:primitiveMatrix}
An irreducible matrix $A$ is primitive when it has only one eigenvalue on the spectral circle. This eigenvalue is equal to the spectral radius of $A$ because the matrix is irreducible. Moreover, when $A$ is primitive, the columns of the powers of $A$ converge to the Perron eigenvector at an exponential rate governed by its second largest eigenvalue (in magnitude) $\lambda_2$.
Specifically,  for any $\zeta$ such that $|\lambda_2| < \zeta < 1$, there exists a positive constant $\kappa$, depending on $A$ and $\zeta$, which guarantees the condition 
\beq 
\left| [A^{i}]_{jk} - v_j \right| \leq \kappa \, \zeta^{i},
\label{eq:matrixPowBound}
\eeq 
for $i=1,2,\ldots$, and for $j,k = 1,2,\ldots,K$.
~\hfill$\square$  
\end{assumption}

\begin{property}[{\bf Sufficient condition for primitivity~\cite{Sayed,sayednewbooks}}]
The combination matrix $A$ is primitive if it is irreducible and has at least one positive element on the main diagonal (i.e., at least one self-loop $a_{kk}>0$). ~\hfill$\square$  
\end{property}

\subsection{Decision Model}
To apply the social learning algorithm \eqref{eq:intSLStep0}-\eqref{eq:socialLearningStep0} we need the decision statistics $d^{\star}_k(x_k;\theta)$, i.e., the log likelihood ratios defined by \eqref{eq:logLikDef}. 
For later use, it is convenient to relate these statistics to the posterior and prior probabilities as follows. 

Given an observation $x_k \in \mathbb{R}^{M_k}$, from Bayes' theorem the \emph{true} posterior probability of $\theta\in\Theta$ is
\beq 
p^{\star}_k(\theta | x_k) \propto \pi^{\star}_k(\theta)\,\ell^{\star}_k(x_k|\theta), 
\label{eq:basicPosterior}
\eeq 
where $\pi^{\star}_k(\theta) > 0$ is the \emph{true} prior probability of hypothesis $\theta$ for agent $k$. 
Now, by introducing the \emph{log posterior ratio}
\beq 
f^{\star}_k(x_k; \theta) \triangleq \log \frac{p^{\star}_k(\theta | x_k)}{p^{\star}_k(\theta_H | x_k)},
\label{eq:logposteriordefinit}
\eeq 
and the {\em log prior ratio}
\beq 
u^{\star}_k(\theta) \triangleq \log \frac{\pi^{\star}_k(\theta)}{\pi^{\star}_k(\theta_H)},
\label{eq:logPriorDef}
\eeq 
from \eqref{eq:basicPosterior} and from the definition of $d^{\star}_k(x_k;\theta)$ in \eqref{eq:logLikDef}, we obtain the relation
\beq 
d^{\star}_k(x_k;\theta) = f^{\star}_k(x_k; \theta) - u^{\star}_k(\theta).
\label{eq:fFuncDef}
\eeq 
In the next section we will show how to estimate $f^{\star}_k(x_k; \theta)$ and $u^{\star}_k(\theta)$ from a training set.
Then, relation \eqref{eq:fFuncDef} will be used to estimate the decision statistic $d^{\star}_k(x_k;\theta)$ from the estimates of $f^{\star}_k(x_k; \theta)$ and $u^{\star}_k(\theta)$.

\section{Model Learning}
\label{sec:SGDExplained}
In most scenarios, the decision models are not available and must be replaced by suitable estimates. 
Under the \emph{supervised learning} framework, these estimates are computed over a training set made of independent and identically distributed (iid) samples $(\bm{x}_{k,t}', \bm{\theta}_{k,t}')$, a.k.a. (feature, label) pairs~\cite{sayednewbooks,VapnikStatLearn}. 
The feature $\bm{x}_{k,t}'$ corresponding to a label $\bm{\theta}_{k,t}'=\theta$ is drawn from the generative model $\ell^{\star}_k(x_k|\theta)$. We recall that the superscript $'$ denotes the {\em training} samples to distinguish them from the {\em prediction} samples $\bm{x}_{k,t}$. Sometimes we omit the subscript $t$ and write $(\bm{x}_{k}', \bm{\theta}_{k}')$ to denote a pair of random variables following the distribution in the training set. Likewise, the notation $\bm{x}_k$ indicates a random variable following the distribution of the prediction data.

We focus on the \emph{discriminative} paradigm, which provides a method to estimate the posterior probabilities $p^{\star}_k(\theta|x_k)$, as illustrated in Sec.~\ref{sec:multiClassSoftmax}. 
Then, in Sec.~\ref{sec:priorML}, we specialize this method to estimate also the prior probabilities $\pi^{\star}_k(\theta)$.
Once the posterior and the prior are estimated, the decision statistic can be obtained by applying \eqref{eq:fFuncDef}.  
This is of course one possibility to learn the decision models from the data. Other strategies might be conceived. 
One motivation for our strategy is that it is based on well-assessed machine learning tools, where the learned functions (i.e., the posterior and the prior probabilities) are chosen so as to minimize interpretable cost functions quantifying the distance with respect to the true data distributions. 
We also remark that the \emph{generative} approach where one would attempt to learn directly the likelihoods $\ell^{\star}_k(x_k|\theta)$ is usually not preferable\footnote{Note that in this work we are estimating log likelihood {\em ratios} and not likelihoods, since we are applying the discriminative approach.}, especially when the features are continuous random variables and/or the feature space is high-dimensional~\cite{NgJordanNIPS2001,VapnikStatLearn}.

\subsection{Learning the Posterior Probabilities}
\label{sec:multiClassSoftmax}
To start with, it is useful to represent the true posterior pmf as a function of the log posterior ratio $f^{\star}_k(x_k; \theta)$ by resorting to the so-called {\em softmax} form~\cite{sayednewbooks,PatternRecognition}:
\beq 
p^{\star}_k(\theta|x_k) = \frac{e^{f^{\star}_k(x_k; \theta)}}{\sum\limits_{\tau \in \Theta} e^{f^{\star}_k(x_k; \tau)}},  
\label{eq:softmaxAppearsFull}
\eeq 
or 
\beq 
p^{\star}_k(\theta | x_k) \propto e^{f^{\star}_k(x_k; \theta)}.
\label{eq:softmaxAppears}
\eeq 
Note that $f^{\star}_k(x_k; \theta_H) = 0$ in view of \eqref{eq:logposteriordefinit}.
Now, in the data-driven framework, the true log posterior ratio $f^{\star}_k(x_k;\theta)$ is replaced by an estimate $f_k(x_k; \theta)$, yielding an estimate $p_k(\theta | x_k)$ of the true posterior $p^{\star}(\theta|x_k)$ in the form
\beq 
p_k(\theta | x_k) \propto e^{f_k(x_k; \theta)}.
\label{eq:softmaxApprox}
\eeq 
The estimate $f_k(x_k; \theta)$ is chosen from a family of admissible functions, and in this work we consider the well-known (multiclass) logistic regression model~\cite{sayednewbooks}:
\beq 
f_{k}(x_k; \theta) = 
\textnormal{col}\{x_k,1\}^{\top}w_{k}(\theta).
\label{eq:fFuncShape}
\eeq 
For notational convenience, from now on we continue to use the same symbol $x_k$ to denote the \emph{extended}\footnote{This is a standard choice to incorporate an offset term, i.e., the last entry of $w_k(\theta)$, into the regression model \eqref{eq:fFuncShape}.} feature vector that incorporates the dummy unit entry. 
Observe that $w_k(\theta)$ is a parameter vector of dimension $M_k+1$. We set $w_k(\theta_H)=0$ to enforce the condition $f_{k}(x_k; \theta_H)=0$.
Using~\eqref{eq:fFuncShape} in the approximate posterior~\eqref{eq:softmaxApprox}, we get:
\beq 
p_k(\theta|x_k) \propto e^{x_k^{\top}w_k( \theta)}.
\label{eq:softmaxAppearsAlternative}
\eeq
It is convenient to aggregate the vectors $w_k(\theta)$, for $\theta\neq\theta_H$, into the extended vector
\beq
w_{k} \! = \! \textnormal{col}\left\{ w_{k}(\theta_1), w_{k}(\theta_2). \ldots, w_{k}(\theta_{H-1})\right\}\!\in\!\mathbb{R}^{(M_k+1)(H-1)}.
\label{eq:parameterVector} 
\eeq
To learn the parameter vector $w_k$, one popular criterion is the minimization of the regularized conditional cross-entropy between the true posterior distribution $p^{\star}_k(\theta|x_k)$ and the approximate posterior $p_k(\theta|x_k)$, which is defined as~\cite{sayednewbooks}
\beq 
J_{k}(w_k) \triangleq -\E \Big[\!\!\!\underbrace{\log p_k(\bm{\theta}_{k}'|\bm{x}_{k}') + \frac{\rho}{2}\|w_k\|^2}_{\textnormal{regularized log-loss $Q_{k}\left(w_k;\bm{x}_{k}',\bm{\theta}_{k}'\right)$}}\!\!\!\Big],
\label{eq:genericCrossEntropy}
\eeq 
where $\rho > 0$ is the regularization parameter and the expectation is computed with respect to the training set distribution.  Note that, on the RHS  of \eqref{eq:genericCrossEntropy}, $p_k(\bm{\theta}_{k}'|\bm{x}_{k}')$ depends implicitly on the unknown vector $w_k$. 

\begin{property}[{\bf Regularized conditional cross-entropy}]
\label{prop:costFuncProp}
The regularized conditional cross-entropy in \eqref{eq:genericCrossEntropy},
with approximate posterior $p_k$ from~\eqref{eq:softmaxAppearsAlternative} is twice differentiable, strongly convex, and has Lipschitz gradient. Namely, for some positive constants $\nu_{k}$ and $\xi_{k}$, it holds that: 
\beq 
\nu_{k} I_{(M_k + 1)(H-1)} \leq \nabla^2 J_{k}(w_k) \leq \xi_{k} I_{(M_k + 1)(H-1)}.
\label{eq:smoothProp}
\eeq 
\end{property}
\begin{IEEEproof}
This result can be proved by extending the arguments required to prove it for the binary logistic regression case~\cite{Sayed, sayednewbooks, PatternRecognition}. 
\end{IEEEproof}

As a consequence of Property~\ref{prop:costFuncProp}, the cost function $J_{k}(w_k)$ has a unique minimizer 
\begin{align}
w_k^{o} 
& = \textnormal{col}\left\{ w^{o}_{k}(\theta_1), w^{o}_{k}(\theta_2) \ldots, w^{o}_{k}(\theta_{H-1})\right\} \triangleq\argmin_{w_k \in \mathbb{R}^{(M_k + 1)(H-1)}} J_{k}(w_k),
\label{eq:wStarDef}
\end{align}
which, however, cannot be computed in closed form since the distribution of the training data $(\bm{x}_{k,t}',\bm{\theta}_{k,t}')$ is unknown. A widely used algorithm to estimate the minimizer $w_k^o$ is the {\em stochastic gradient descent} (SGD) algorithm, which, when equipped with a {\em constant step-size}, also guarantees the critical adaptation capabilities that are required to act in an online setting~\cite{Sayed}. The SGD algorithm is defined by the recursion 
\beq 
\bm{w}_{k,t} = \bm{w}_{k,t-1} - \eta \, \nabla Q_{k}\left(\bm{w}_{k,t-1};\bm{x}_{k,t}',\bm{\theta}_{k,t}'\right),
\label{eq:asynBasicSGDRecursionW}
\eeq 
where the gradient is taken with respect to the first argument of the loss function and $\eta > 0$ is the step-size that controls the degree of adaptation. 
Since we are now dealing with the training phase, we will refer to $\eta$ as the \emph{training adaptation parameter}. 
For sufficiently large $t$ and sufficiently small step-size $\eta$, the SGD recursion approximates, up to a constant error floor, the unique minimizer $w_k^o$, in the sense that~\cite{Sayed,sayednewbooks},
\beq
\limsup_{t\rightarrow\infty}\E\|\bm{w}_{k,t} - w_k^{o}\|^2=O(\eta).
\label{eq:MSDclassic}
\eeq
By substituting $\bm{w}_{k,t}$ into \eqref{eq:fFuncShape}, we obtain the following online estimate of the log posterior ratio, for $\theta\neq\theta_H$:
\beq 
\bm{f}_{k,t}(x_k; \theta) = x_{k}^{\top}\bm{w}_{k,t}(\theta).
\label{eq:onlinelogposterior}
\eeq

\subsection{Learning the Prior Probabilities}
\label{sec:priorML}
The approach illustrated in the previous section can be simplified to estimate the prior probabilities. 
First, we use the softmax form to represent the estimated prior probabilities $\pi_k(\theta)$ in terms of the estimated log prior ratios $u_k(\theta)$, yielding
\beq 
\pi_k(\theta) \propto e^{u_k(\theta)},
\label{eq:softmaxAppearsAlternativePi}
\eeq
where $u_{k}(\theta) \in \mathbb{R}$.
Note that, while the posterior depends on $x_k$, the prior depends only on $\theta$. In other words, we will only need the labels to estimate the prior.
It is also convenient to aggregate the parameters $u_k(\theta)$, for $\theta\neq\theta_H$, into the vector
\beq 
u_k = \textnormal{col}\{ u_k(\theta_1), u_k(\theta_2), \ldots, u_k(\theta_{H-1}) \} \in \mathbb{R}^{H-1}.
\eeq 
Following the same approach used in the previous section, the parameters $u_k$ can be learned by minimizing the cross-entropy between the true prior distribution $\pi^{\star}_k(\theta)$ and the approximate prior distribution $\pi_k(\theta)$, yielding (we will use the symbol $\widetilde{\phantom{x}}$ to denote all quantities related to the estimation of the prior)
\beq 
\widetilde{J}_{k}(u_k) \triangleq -\E \Big[\!\!\!\underbrace{\log \pi_k(\bm{\theta}_{k}') + \frac{\widetilde{\rho}}{2}\|u_k\|^2}_{\textnormal{regularized log-loss $\widetilde{Q}_{k}\left(u_k;\bm{\theta}_{k}'\right)$}}\!\!\!\Big],
\label{eq:genericCrossEntropyU}
\eeq 
where $\widetilde{\rho} > 0$ is the regularization parameter. Property~\ref{prop:costFuncProp} also holds for the cost function $\widetilde{J}_{k}(u_k)$, with suitable constants $\widetilde{\nu}_{k}$ and $\widetilde{\xi}_{k}$, which implies that $\widetilde{J}_{k}(u_k)$ has a unique minimizer 
\begin{align}
u_k^{o} & = \textnormal{col}\left\{ u^{o}_{k}(\theta_1), \ldots, u^{o}_{k}(\theta_{H-1})\right\} \triangleq \argmin_{u_k \in \mathbb{R}^{(H-1)}} \widetilde{J}_{k}(u_k).
\label{eq:uStarDef}
\end{align} 
The minimizer $u_k^{o}$ can be estimated by means of the following SGD recursion, 
\beq 
\bm{u}_{k,t} = \bm{u}_{k,t-1} - \widetilde{\eta} \, \nabla \widetilde{Q}_{k}\left(\bm{u}_{k,t-1};\bm{\theta}_{k,t}'\right),
\label{eq:asynBasicSGDRecursionU}
\eeq 
where the gradient is computed with respect to the first argument of $\widetilde{Q}_k$.
The iterates $\bm{u}_{k,t}$ approximate $u_k^{o}$ in the same sense as~\eqref{eq:MSDclassic}, namely,
\beq
\limsup_{t\rightarrow\infty}\E\|\bm{u}_{k,t} - u_k^{o}\|^2=O(\widetilde{\eta}).
\label{eq:MSDclassic2}
\eeq

\subsection{Learning the Decision Statistics}

Consider now \eqref{eq:fFuncDef}. Replacing the true log posterior ratio $f^{\star}_k(x_k;\theta)$ with the online estimate $\bm{f}_{k,t}(x_k;\theta)$ from \eqref{eq:onlinelogposterior}, and the true log prior ratio $u^{\star}_k(\theta)$ with the online estimate $\bm{u}_{k,t}(\theta)$ from \eqref{eq:asynBasicSGDRecursionU}, we obtain the following online estimate for the log likelihood ratio $d^{\star}_k(x_k;\theta)$:
\beq 
\bm{d}_{k,t}(x_k;\theta) = \bm{f}_{k,t}(x_k;\theta) -  \bm{u}_{k,t}(\theta) =
x_k^{\top}\bm{w}_{k,t}(\theta) - \bm{u}_{k,t}(\theta).
\label{eq:finalDecStat}
\eeq
We remark that $\bm{d}_{k,t}$ is a \emph{random function} (bold font) because it results from an optimization process based on random training samples. In other words, the randomness of $\bm{d}_{k,t}$ stems from the randomness of the training set. 
Later on, when we evaluate $\bm{d}_{k,t}$ at specific \emph{prediction} samples $\bm{x}_{k,t}$, the randomness of the prediction stage will also be involved.

\section{$\asl$ Strategy}
The doubly adaptive social learning strategy proposed in this work consists of the operations listed in Algorithm~\ref{alg:a2sl}, performed iteratively by all agents at each time instant $t \geq 1$ (at time $t = 0$, the agents are initialized with deterministic vectors $w_{k,0}$ and $u_{k,0}$):

\begin{algorithm} 
\caption{$\asl$} 
\label{alg:a2sl} 
\small
\renewcommand{\algorithmicindent}{0.5em}
\hspace*{\algorithmicindent} 
\begin{algorithmic}[1]
\FOR{$t=1,2,\ldots$}
	\STATE \textbf{for} $k=1,2,\ldots,K$ \textbf{do}
	\begin{subequations}
	\begin{align}
	& \bm{w}_{k,t} = \bm{w}_{k,t-1} - \eta \nabla Q_{k}\!\left(\bm{w}_{k,t-1}; \bm{x}_{k,t}',\bm{\theta}_{k,t}'\right)\bm{\alpha}^{\rm{tr}}_{k,t},
	\label{eq:trainStep1} \\
	& \bm{u}_{k,t} = \bm{u}_{k,t-1} - \widetilde{\eta}\,\nabla \widetilde{Q}_{k}\left(\bm{u}_{k,t-1}; \bm{\theta}_{k,t}'\right)\bm{\alpha}^{\rm{tr}}_{k,t},
	\label{eq:trainStep2} \\
	& \bm{d}_{k,t}(\bm{x}_{k,t}; \theta) = 
	\bm{x}_{k,t}^{\top}\bm{w}_{k,t}(\theta) - \bm{u}_{k,t}(\theta),
	\label{eq:decModelUpdate}
	\\
	& \bm{\psi}_{k,t}(\theta) \propto\bm{\mu}_{k,t-1}^{1-\delta}(\theta)\exp\left\{\bm{d}_{k,t}(\bm{x}_{k,t};\theta)\,\bm{\alpha}^{\rm{pr}}_{k,t}\right\}
	\label{eq:intSLStepApprox}, \\
	& \bm{\mu}_{k,t}(\theta) \propto \prod_{j=1}^K[\bm{\psi}_{j,t}(\theta)]^{a_{jk}}.
	\label{eq:socialLearningStep}
	\end{align}
	\end{subequations}
    \STATE \textbf{end for}
\ENDFOR
\end{algorithmic}
\end{algorithm}

\noindent
-- {\em Model Parameter Update}. In \eqref{eq:trainStep1}, each agent $k$ performs an SGD iteration to update the parameter vector $\bm{w}_{k,t}$ after observing the training sample $(\bm{x}_{k,t}', \bm{\theta}_{k,t}')$. In comparison to \eqref{eq:asynBasicSGDRecursionW}, we added the scalar $\bm{\alpha}^{\rm{tr}}_{k,t}$, which is assumed to be a Bernoulli random variable taking value $1$ with probability $q_k^{\rm{tr}}>0$. 
When $\bm{\alpha}^{\rm{tr}}_{k,t} = 0$, no training sample at time $t$ is observed and the parameter vector is not updated. The insertion of this variable allows for a flexible framework, where training and prediction phases can take place in an asynchronous manner. At any time instant, the agents can observe a new training sample, and/or a new prediction sample, or no samples at all. 
\\
-- {\em Prior Probabilities Update}. In \eqref{eq:trainStep2}, each agent $k$ updates with an SGD iteration the parameter vector $\bm{u}_{k,t}$ employed to approximate the priors. Similarly to the model parameter update, this SGD iteration is performed only if a new training sample is observed, according to the probability law enforced by the Bernoulli random variable $\bm{\alpha}^{\rm{tr}}_{k,t}$. \\
-- {\em Decision Model Computation}. In \eqref{eq:decModelUpdate}, each agent $k$ computes the current decision model by using the updated vectors $\bm{w}_{k,t}$ and $\bm{u}_{k,t}$. 
\\
-- {\em Adaptive Bayesian Update}. In \eqref{eq:intSLStepApprox}, the agents update their intermediate beliefs according to the adaptive rule proposed in~\cite{bordignon2021adaptive}, using the adaptation parameter $\delta\in (0,1)$. The role of $\delta$ is to discount the past beliefs. In this way, the new observations are given more importance, speeding up the reaction to hypothesis drifts. Since $\delta$ controls the adaptation degree in the prediction phase, we will refer to it as the \emph{prediction adaptation parameter}. 
For the same reasons discussed in relation to the training samples, we assume that agent $k$ observes prediction samples $\bm{x}_{k,t}$ with probability $q_k^{\rm{pr}}>0$. Note that if no prediction samples are observed, the likelihood is not informative, therefore in \eqref{eq:intSLStepApprox} we introduce as a multiplicative factor the Bernoulli random variable  $\bm{\alpha}^{\rm{pr}}_{k,t}$.
\\ 
-- {\em Combination Step}. In \eqref{eq:socialLearningStep}, the agents compute their beliefs by aggregating the intermediate updates received from their neighbors.

It is convenient to introduce the log belief ratios:
\beq 
\bm{\beta}_{k,t}(\theta) \triangleq \log \frac{\bm{\mu}_{k,t}(\theta_0)}{\bm{\mu}_{k,t}(\theta)},
\label{eq:logBelDef}
\eeq 
where $\theta_0$ denotes the true hypothesis giving rise to the \emph{prediction} data.
From \eqref{eq:intSLStepApprox} and \eqref{eq:socialLearningStep}, we can evaluate the evolution of the log belief ratios at agent $k$, for $\theta \neq \theta_0$, with the recursion
\beq 
\bm{\beta}_{k,t}(\theta) = \sum_{j = 1}^K a_{j k} \left\{ (1-\delta)\bm{\beta}_{j,t-1}(\theta) +  \bm{\lambda}_{j,t}(\theta) \right\},
\label{eq:logBelRecurs}
\eeq 
where we introduced the quantity:
\begin{align} 
\bm{\lambda}_{k,t}(\theta) & \triangleq \bm{\alpha}^{\rm{pr}}_{k,t}\left[\bm{d}_{k,t}(\bm{x}_{k,t} ; \theta_0) - \bm{d}_{k,t}(\bm{x}_{k,t} ; \theta) \right] 
\nonumber \\
& = \bm{\alpha}^{\rm{pr}}_{k,t}\bm{x}_{k,t}^{\top}\big[\bm{w}_{k,t}(\theta_0) - \bm{w}_{k,t}(\theta) \big] - \bm{\alpha}^{\rm{pr}}_{k,t}\left[ \bm{u}_{k,t}(\theta_0) - \bm{u}_{k,t}(\theta)\right] \nonumber \\
& = \bm{\alpha}^{\rm{pr}}_{k,t}
\left[
\bm{x}_{k,t}^{\top}\bm{\Delta}_{k,t}(\theta) - \widetilde{\bm{\Delta }}_{k,t}(\theta)
\right],
\label{eq:logLikDef2}
\end{align}
with
\begin{align} 
\bm{\Delta}_{k,t}(\theta) &\triangleq \bm{w}_{k,t}(\theta_0) - \bm{w}_{k,t}(\theta), \nonumber\\ \widetilde{\bm{\Delta}}_{k,t}(\theta) &\triangleq \bm{u}_{k,t}(\theta_0) - \bm{u}_{k,t}(\theta).
\label{eq:deltaWdef}
\end{align} 
The recursion in \eqref{eq:logBelRecurs} can be unfolded to obtain
\begingroup
\allowdisplaybreaks
\begin{align}
& \bm{\beta}_{k,t}(\theta) = \underbrace{ (1-\delta)^t\sum_{j = 1}^K  [A^t]_{j k}\bm{\beta}_{j,0}(\theta)}_{\bm{\gamma}_t} + \sum_{m=0}^{t-1} \sum_{j = 1}^K (1-\delta)^m [A^{m+1}]_{j k}\bm{\lambda}_{j, t-m}(\theta)  \nonumber \\
& = \bm{\gamma}_t + \sum_{m=0}^{t-1} \sum_{j = 1}^K (1-\delta)^m [A^{m+1}]_{j k} \bm{\alpha}^{\rm{pr}}_{j,t-m} \left(\bm{x}_{j,t-m}^{\top}\bm{\Delta}_{j, t-m}(\theta) -  \widetilde{\bm{\Delta}}_{j,t-m}(\theta)\right) ,
\label{eq:slRecDev}
\end{align} 
\endgroup
where the second equality follows from \eqref{eq:logLikDef2}. 

\section{$\asl$ Learning Behavior}
The learning behavior of the $\asl$ strategy will be examined resorting to powerful tools from the theory of adaptation and learning~\cite{bordignon2021adaptive, Sayed, sayednewbooks}. We will consider an arbitrary time instant $t_0$, and some drift in the data and/or models which will be effective from $t_0+1$ onward. The initial beliefs $\mu_{k,t_0}$ and parameters $w_{k,t_0}$, $u_{k,t_0}$ represent, for a fixed realization of the training and prediction data, the only quantities required to execute the algorithm \eqref{eq:trainStep1}--\eqref{eq:socialLearningStep} from $t_0 + 1$ onward, and store all the past knowledge on the inferential problem accumulated by the agents up to $t_0$. Our analysis characterizes the system evolution over a stationary interval, i.e, an interval where no drifts happen, starting from $t_0+1$. For notation convenience, and without loss of generality, we set $t_0 = 0$.

\begin{assumption}[{\bf Prediction-data properties}]
\label{ass:trainingBernoulli}
For each agent $k$, the observations $\bm{x}_{k,t}$ are iid random variables with finite second moment, generated from the model $\ell^{\star}_k(x_k|\theta_0)$, where $\theta_0$ is the actual hypothesis in force from $t_0$\footnote{We remark that the observations are allowed to be dependent across the agents. Nevertheless, we do not need to specify the joint dependence, since social learning enables faithful learning by considering only marginal models, i.e., models not encompassing joint dependence across the agents.}.
Prediction and training data are independent, and the Bernoulli random variables $\bm{\alpha}^{\rm{tr}}_{k,t}$ and $\bm{\alpha}^{\rm{pr}}_{k,t}$ are iid over time and across the agents, mutually independent, as well as independent from the training and prediction data. 
\end{assumption}

We will characterize the learning behavior of the $\asl$ strategy by examining: $(i)$ the transient phase, where we establish that the algorithm adapts to changing conditions at an exponential rate; and $(ii)$ the steady-state phase, where we establish that, as the hypothesis remains stable for sufficiently long time, the algorithm identifies the true hypothesis with a small error probability on the order of the adaptation parameters.
First of all, we need to introduce the performance measure that quantifies the learning capabilities of social learning strategies. Every agent in the network makes, for each time instant $t$, its decision on the true state of nature by choosing the hypothesis that maximizes its belief vector $\bm{\mu}_{k,t}$. The performance will be accordingly measured by
the instantaneous error probability of agent $k$:
\begin{align}
p_{k,t} \triangleq \mathbb{P}\Big[ \theta_0 \neq \argmax_{\theta \in \Theta} \bm{\mu}_{k,t}(\theta) \Big] \!=\! \mathbb{P}\Big[ \exists \theta \neq \theta_0 \!:\! \bm{\beta}_{k,t}(\theta) \!\leq\! 0 \Big].
\label{eq:errorProbDef}
\end{align}
We will prove that, for sufficiently small adaptation parameters, the instantaneous error probability $p_{k,t}$ decays {\em exponentially fast}, converging to a small value on the order of the adaptation parameters. In other words, in the steady-state, the error probabilities of all agents vanish as the adaptation parameters vanish. For this purpose, we will resort to a global identifiability condition that extends to the unknown likelihood setting the traditional identifiability condition adopted in traditional social learning. 

To clarify the rationale behind the identifiability condition that we will use, let us introduce the optimal decision statistic $d_k^{o}(x_k;\theta)$ derived from \eqref{eq:finalDecStat} by employing the optimal parameters $w^{o}_k$ and the optimal log prior ratios $u_k^{o}$.
Let  
\beq 
\Delta_k^{o}(\theta) \triangleq w^{o}_k(\theta_0) -  w^{o}_k(\theta), \quad 
\widetilde{\Delta}_k^{o}(\theta) \triangleq u^{o}_k(\theta_0) -  u^{o}_k(\theta),
\label{eq:deltaOptimalDef}
\eeq 
be the quantities in~\eqref{eq:deltaWdef} computed using the optimal parameter vectors $w^{o}_k$ and $u^{o}_k$.

Consider a single-agent (SA) version of \eqref{eq:intSLStep0}--\eqref{eq:socialLearningStep0} (i.e., ignore the combination step and set $\bm{\psi}^{\rm{SA}}_{k,t}=\bm{\mu}^{\rm{SA}}_{k,t}$) and use $d_k^{o}(x_k;\theta)$ in place of $d^{\star}_k(x_k;\theta)$, obtaining:
\beq
\bm{\mu}^{\rm{SA}}_{k,t}(\theta)\!\propto\! \bm{\mu}^{\rm{SA}}_{k,t-1}(\theta)e^{d_k^{o}(\bm{x}_{k,t};\theta)}\!=\!\mu^{\rm{SA}}_{k,0}(\theta)e^{\sum_{m=1}^t \!d_k^{o}(\bm{x}_{k,m};\theta)}.
\eeq
By the law of large numbers we get:
\beq
\lim_{t\rightarrow\infty}\frac 1 t\log \frac{\bm{\mu}^{\rm{SA}}_{k,t}(\theta_0)}{\bm{\mu}^{\rm{SA}}_{k,t}(\theta)}
=
\E\left[d_k^{o}(\bm{x}_{k,t};\theta_0) - d_k^{o}(\bm{x}_{k,t};\theta) \right],
\label{eq:preSSLNagent}
\eeq
with probability $1$.
Now, if\hspace{1pt}\footnote{We note in passing that when the likelihood ratios are perfectly known, i.e., when $d^{o}_{k}(x_k;\theta) = d^{\star}_{k}(x_k;\theta) =\log \left( \ell^{\star}_{k}(x_k|\theta)/ \ell^{\star}_{k}(x_k|\theta_H) \right)$, condition \eqref{eq:localident} corresponds to requiring that the KL divergence between $\ell^{\star}_{k}(x_k|\theta_0)$ and $\ell^{\star}_{k}(x_k|\theta)$ is strictly positive, i.e., that any $\theta$ is distinguishable from $\theta_0$ or, equivalently, that $\theta_0$ is identifiable.}
\begin{equation}
\E\!\left[d_k^{o}(\bm{x}_{k,t};\theta_0)
\!-\!
d_k^{o}(\bm{x}_{k,t};\theta)\right] \!=\!
\E\,\bm{x}_{k}^{\top} \Delta_k^{o}(\theta)  - \widetilde{\Delta}_k^{o}(\theta) > 0,
\label{eq:localident}
\end{equation}
from \eqref{eq:preSSLNagent} we conclude that the true hypothesis can be perfectly identified by agent $k$ over an infinite stream of data. For this reason, we refer to \eqref{eq:localident} as a {\em local identifiability} condition. In our {\em social} learning framework, cooperation among agents allows us to {\em relax} the local identifiability at each individual agent with the following less stringent condition, which takes into account {\em all} the agents involved in the $\asl$ strategy. 
\begin{assumption}[{\bf Global identifiability}]
\label{ass:globalIdent}
We say that global identifiability holds when, for all pairs $(\theta,\theta_0)$ with $\theta\neq\theta_0$:
\begin{align}
\beta_{\rm{net}}(\theta) & \triangleq  
\sum_{j=1}^K v_{k} \, q_j^{\rm{pr}}
\left(\E\,\bm{x}_{j}^{\top}\Delta_j^{o}(\theta) - \widetilde{\Delta}_j^{o}(\theta) \right)
>0.
\label{eq:limitLogBel}
\end{align}
\end{assumption}
 
We can get an interpretation of condition \eqref{eq:limitLogBel} in connection with the local identifiability in \eqref{eq:localident}, by introducing the aggregate, across the agents, decision model:
\beq
\sum_{j=1}^K v_{j} \, \bm{\alpha}_{j,t}^{\rm{pr}} \,d_j^{o}(\bm{x}_{j,t};\theta).
\label{eq:aggrStatNew}
\eeq
Condition \eqref{eq:limitLogBel} can now be interpreted as the counterpart of \eqref{eq:localident} where the local decision statistic $d_k^{o}(\bm{x}_{k,t};\theta)$ is replaced by the aggregate statistic \eqref{eq:aggrStatNew}.
The condition $\beta_{\rm{net}}(\theta)>0$ is less stringent than local identifiability because it does not require that the individual terms of the summation in \eqref{eq:limitLogBel} are positive for all $k$, i.e., that local identifiability holds for all agents, but rather their combination.
It is important to highlight that the role of each agent in the identifiability condition is magnified by its local attributes. In particular, agents with higher network centrality, i.e., characterized by high values of the Perron eigenvector entries $v_k$, and dense data acquisitions, i.e., characterized by higher probability $q_k^{\rm{pr}}$, are more influential in assessing the sign of the limit value $\beta_{\rm{net}}(\theta)$.

\begin{theorem}[{\bf $\asl$ consistency}] 
\label{th:consistency}
Assume positive deterministic initial beliefs $\mu_{k,0}(\theta)$ for all the agents and all the hypotheses. For $k=1,2,\ldots,K$, let
\beq 
\delta < q_k^{\rm{tr}} \left(
\min \left\{ \frac{\nu_{k}}{\xi_{k}}, \frac{\widetilde{\nu}_{k}}{\widetilde{\xi}_{k}}
 \right\}\right)^2,
\label{eq:deltaCond}
\eeq 
and
\beq
\eta < \min\left\{\frac{2\,\nu_{k}}{\xi_{k}^2}, \eta^{o}\right\}, \; 
\widetilde{\eta} < \min\left\{\frac{2\,\widetilde{\nu}_{k}}{\widetilde{\xi}_{k}^2}, \widetilde{\eta}^{o}\right\},
\label{eq:stabConditionEta}
\eeq 
where $\eta^{o}$ and $\widetilde{\eta}^{o}$ are the smallest roots of the equations\footnote{Under condition \eqref{eq:deltaCond} both equations have two positive roots.} 
\beq 
q_k^{\rm{tr}}
\left(
\eta^2_{k}\,\xi^2_{k} - 2\eta_{k}\,\nu_{k}
\right) +\delta = 0,
\eeq
\beq 
q_k^{\rm{tr}}\left(
\widetilde{\eta}^2_{k}\,\widetilde{\xi}^2_{k} -2\widetilde{\eta}_{k}\,\widetilde{\nu}_{k}\right) +\delta = 0.
\eeq 
Then, the SGD recursions \eqref{eq:trainStep1} and \eqref{eq:trainStep2} estimate $w_k^{o}$ and $u_k^{o}$, respectively, by reaching the error floors \eqref{eq:MSDclassic} and \eqref{eq:MSDclassic2} according to the laws
\beq 
\E\| \bm{w}_{k, t} - w_k^{o} \|^2 \leq \varphi_{k}^t \E\| \bm{w}_{k,0} - w_k^{o} \|^2 + O(\eta),
\label{eq:sgdMSERecursionWTheorem}
\eeq
\beq 
\E\| \bm{u}_{k, t} - u_k^{o} \|^2 \leq \widetilde{\varphi}_{k}^t \E\| \bm{u}_{k,0} - u_k^{o} \|^2 + O(\widetilde{\eta}),
\label{eq:sgdMSERecursionUTheorem}
\eeq
with convergence rates $\varphi_{k}$ and $\widetilde{\varphi}_{k}$ taking values in $(0,1]$ and being equal to
\begin{align}
& \varphi_{k} = 1 - 2\,\eta\, q_k^{\rm{tr}}\nu_{k} + \eta^2\,q_k^{\rm{tr}}\xi_{k}^2, 
\label{eq:convergenceRateW}
\\
& \widetilde{\varphi}_{k} = 1 - 2\,\widetilde{\eta}\, q_k^{\rm{tr}}\widetilde{\nu}_{k} + \widetilde{\eta}^2\,q_k^{\rm{tr}}\widetilde{\xi}_{k}^2. 
\label{eq:convergenceRateU1}
\end{align}
Let
\beq 
\varphi \triangleq \max_{k=1,2,\ldots,N} \varphi_{k}, \quad \widetilde{\varphi} \triangleq \max_{k=1,2,\ldots,N} \widetilde{\varphi}_{k}.
\label{eq:slowestConvRates}
\eeq
Then, under Assumptions~\ref{ass:irreducibleMatrix}-\ref{ass:globalIdent} the instantaneous error probability of each agent evolves according to the bound
\begin{align} 
p_{k,t} & \leq \underbrace{
\delta\left(
c_1 \,\varphi^t  + 
c_2\,\widetilde{\varphi}^t 
\right)}_{\textnormal{model-learning transient}} + 
\underbrace{
c_3 \, (1-\delta)^{2t} }_{\textnormal{social-learning transient}} + \underbrace{c_4\,\delta + c_5\,\eta + 
c_6\,\widetilde{\eta}}_{\textnormal{steady-state term}},
\label{eq:OSLConsistency}
\end{align}
where the expressions for the constants $c_1, c_2,\ldots,c_6$, are detailed in Appendix~\ref{app:consistencyProof}.
\end{theorem}
\begin{IEEEproof}
See Appendix~\ref{app:consistencyProof}.
\end{IEEEproof}

The claim of Theorem~\ref{th:consistency} completely characterizes the learning behavior of the $\asl$ strategy, accounting  for both the transient phase required to react to a drift and the steady state attained after the transient phase. Whenever a drift is met, a new stationary learning cycle starts from the initial conditions met at the end of the previous cycle, i.e., the belief vectors at the end of the previous cycle.
For sufficiently small values of the adaptation parameters $\delta$, $\eta$, and $\widetilde{\eta}$, the error probability of each agent diminishes at a geometric rate and reaches a small steady-state value, i.e., the $\asl$ strategy adapts by promptly tracking the changes in the models and/or hypotheses.
This goal is attained by jointly exploiting the adaptation capabilities of the model-learning stage (SGD with constant step-size)~\cite{Sayed} and of the adaptive social learning stage (adaptive Bayesian update)~\cite{bordignon2021adaptive} that characterize the $\asl$ strategy. 

As time progresses, when the transient terms in \eqref{eq:OSLConsistency} are annihilated, the error probability bound assumes a constant value depending on the adaptation parameters $\delta$, $\eta$ and $\widetilde{\eta}$, as stated by the following Corollary associated with Theorem~\ref{th:consistency}.

\begin{corollary}[{\bf $\asl$ steady-state performance}]
\label{cor:asymPerf}
As $t \rightarrow \infty$ each agent $k$ detects the correct hypothesis $\theta_0$ with an instantaneous error probability on the order of the adaptation parameters $\delta, \eta, \widetilde{\eta}$, namely,
\beq 
\limsup_{t \rightarrow \infty} p_{k,t} = O(\delta) + O(\eta) + O(\widetilde{\eta}).
\label{eq:OSLAsymConsistency}
\eeq 
\end{corollary}
\begin{IEEEproof}
The corollary follows from Theorem~\ref{th:consistency} by applying the limit superior as $t\rightarrow \infty$ to \eqref{eq:OSLConsistency}. 
\end{IEEEproof}

In view of Corollary~\ref{cor:asymPerf}, in the steady state, the agents learn increasingly well as smaller values of the adaptation parameters are chosen. Accordingly, as $t \rightarrow \infty$, and in the regime $\delta, \eta, \widetilde{\eta} \rightarrow 0$, the instantaneous error probability of each agent becomes negligible, and each agent correctly identifies the true hypothesis $\theta_0$. 

\begin{figure*}[t]
\centering
\begin{minipage}{.49\textwidth}
\begin{subfigure}{\textwidth}
\includegraphics[width=\textwidth]{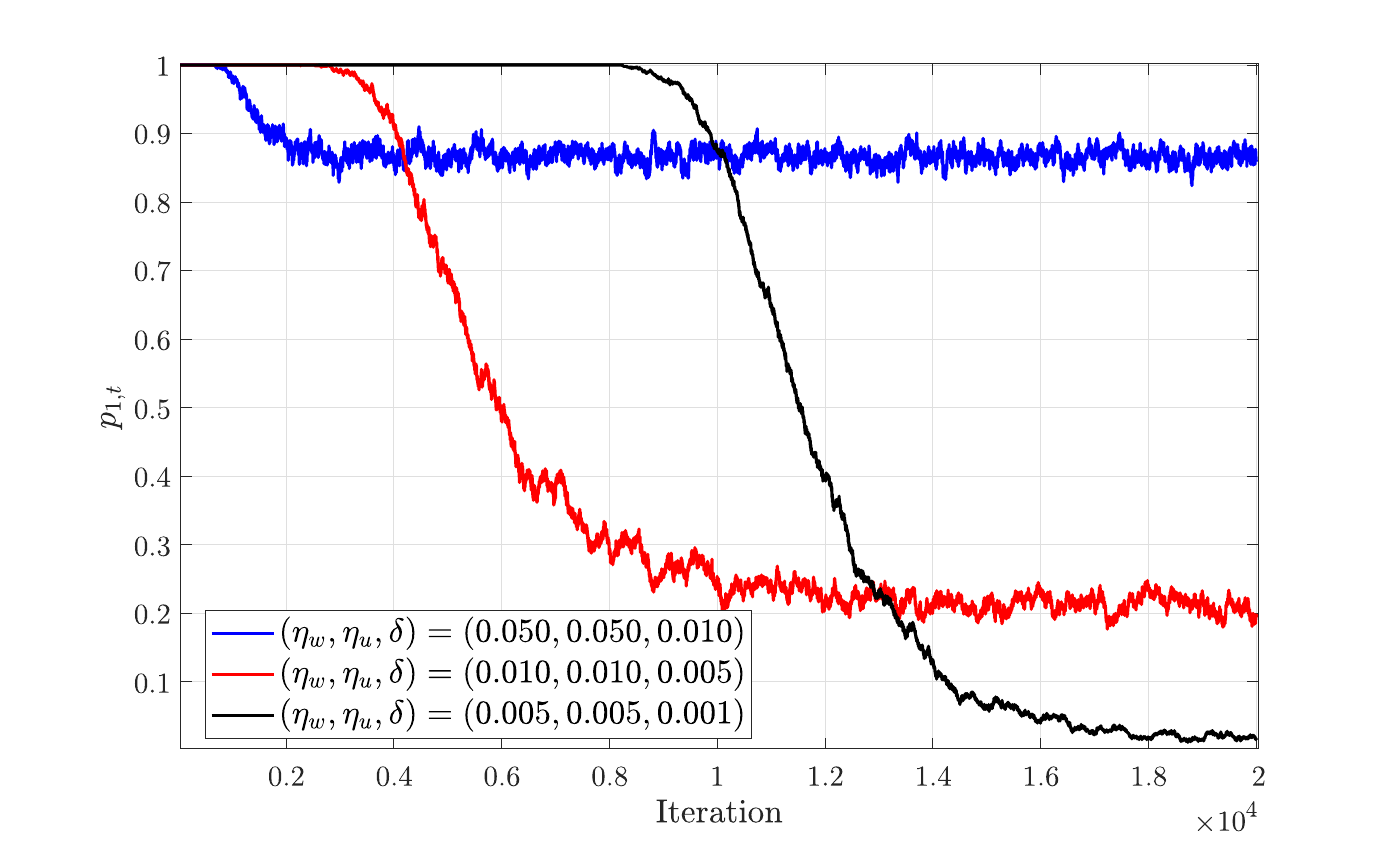}
\end{subfigure}
\end{minipage}
\begin{minipage}{0.49\textwidth}
\begin{subfigure}{\textwidth}
\includegraphics[width=\textwidth]{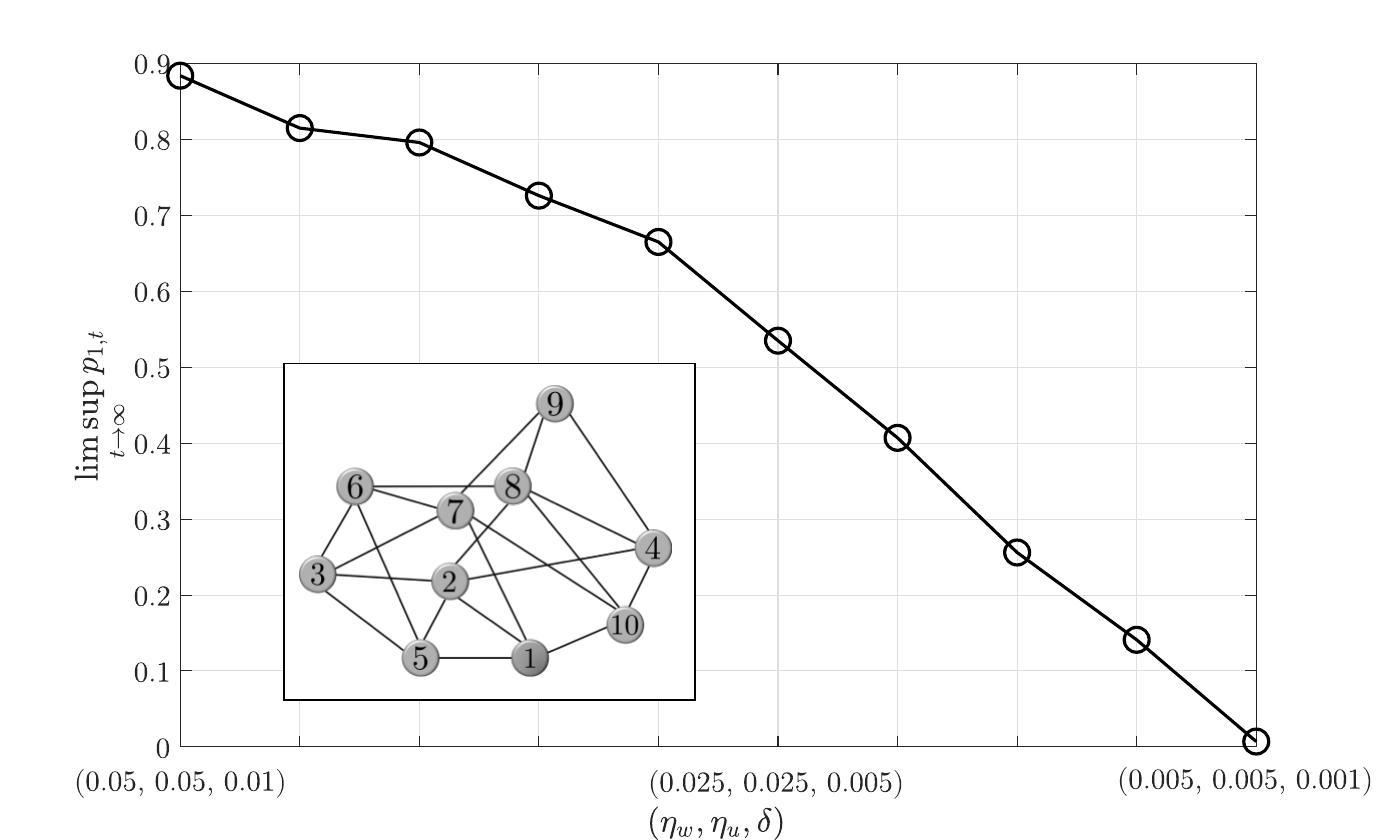}
\end{subfigure}
\end{minipage}
\caption{
{\em Left}. Error probability of agent $1$ for different values of the adaptation parameters. The regularization parameters are set as $\rho = 0.05$, $\widetilde{\rho} = 0.05$, and $q_k^{\rm{pr}} = 0.8$ and $q_k^{\rm{tr}} = 0.7$ for all $k$. The limit value $\beta_{\rm{net}}(\theta)$ to verify the global identifiability condition was computed by evaluating offline the optimal parameters $w_k^{o}$ and $u_k^{o}$ using a stochastic gradient descent with decaying step-size and batch-size equal to $2000$. The network topology is shown in the inset plot of the right panel (all agents have self-loop not shown for ease of illustration). The combination matrix is obtained through the uniform-averaging rule~\cite{Sayed, sayednewbooks} and can be verified to be primitive.
{\em Right}. Error probability of agent $1$ in steady-state conditions for several triplets of adaptation parameters, where the values of $\eta$ and $\widetilde{\eta}$ are uniformly spaced in the interval $[0.005, 0.05]$ and the values of $\delta$ are uniformly spaced in the interval $[0.001, 0.01]$. All curves are estimated by means of $10^3$ Monte Carlo runs. For all plots, similar behavior is observed for the other agents.}
\label{fig:fig1}
\end{figure*}

The results from Theorem~\ref{th:consistency} and Corollary~\ref{cor:asymPerf} exhibit the fundamental trade-off arising in the theory of adaptation and learning~\cite{Sayed, sayednewbooks}. While smaller adaptation parameters ensure better learning performance, i.e., smaller $p_{k,t}$ as $t \rightarrow \infty$, they also increase the convergence rates $\varphi$, $\widetilde{\varphi}$ and $1-\delta$ in \eqref{eq:OSLConsistency}, slowing down the learning process. We will show some examples of the learning-adaptation trade-off in the next section.

\section{Illustrative Examples}

\subsection{Synthetic Data}
\label{sec:exampleA}
The first example illustrates the theoretical findings from Theorem~\ref{th:consistency} and Corollary~\ref{cor:asymPerf}. 
Consider a network of $K=10$ agents connected as shown in the inset plot of the right panel in Fig.~\ref{fig:fig1}, with edges weighted according to the uniform-averaging combination policy~\cite{Sayed, sayednewbooks}. Each agent $k$ observes training samples $(\bm{x}_{k,t}', \bm{\theta}_{k,t}')$, where $\bm{\theta}_{k,t}'$ is sampled from the set $\Theta = \{ 1, 2, 3 \}$ with prior probabilities $[0.3, 0.4, 0.3]$, and the features $\bm{x}_{k,t}'$ are $M_k$-dimensional Gaussian vectors, with $M_k=4$. 
For $m=1,2,\ldots,M_k$, the $m$-th component of the feature vector, denoted by $\bm{x}'_{k,t}(m)$, is generated according to the density function
\beq 
g_n\left(x'_{k,t}(m)\right) = \frac{1}{0.1\,k\sqrt{2\pi}} \exp\left[-0.5\left(\frac{x'_{k,t}(m) - n }{0.1\,n}\right)^2\right],
\eeq 
with $n=1,2,3$. 

Regarding the prediction data, we assume that the true hypothesis is $\theta_0 = 1$. 
In other words, the prediction samples are $M_k$-dimensional Gaussian vectors whose components $\bm{x}_{k,t}(m)$ are distributed according to the pdf $g_1$. 

Cooperation through social learning is crucial in this example, since the agents face a locally non-identifiable problem as detailed in Table~\ref{tab:confoundingMatrix}. Nevertheless, it is readily verified that the global identifiability condition \eqref{eq:limitLogBel} is met.

\begin{table}[h]
\renewcommand{\arraystretch}{1.25}
\centering
\caption{Local non-identifiability setup for the examples in Fig.~\ref{fig:fig1}.}
\begin{tabular}{|p{1cm}p{1cm}|p{1cm}p{1cm}p{1cm}|}
\hline
\multicolumn{2}{|c|}{\multirow{2}{*}{\textbf{Agent $k$}}} & \multicolumn{3}{c|}{\textbf{True likelihood models $\ell^\star_k(\cdot|\theta)$} }                            \\ \cline{3-5} 
\multicolumn{2}{|c|}{}                  & \multicolumn{1}{c|}{$\theta = 1$} & \multicolumn{1}{c|}{$\theta = 2$} & \multicolumn{1}{c|}{$\theta = 3$} \\ \hline
\multicolumn{2}{|c|}{1}                & \multicolumn{1}{c|}{$g_1(\cdot)$} & \multicolumn{1}{c|}{$g_2(\cdot)$} & \multicolumn{1}{c|}{$g_3(\cdot)$} \\ \hline
\multicolumn{2}{|c|}{2-3}              & \multicolumn{1}{c|}{$g_1(\cdot)$} & \multicolumn{1}{c|}{$g_2(\cdot)$} & \multicolumn{1}{c|}{$g_2(\cdot)$} \\ \hline
\multicolumn{2}{|c|}{4-6}              & \multicolumn{1}{c|}{$g_1(\cdot)$} & \multicolumn{1}{c|}{$g_1(\cdot)$} & \multicolumn{1}{c|}{$g_3(\cdot)$} \\ \hline
\multicolumn{2}{|c|}{7-10}             & \multicolumn{1}{c|}{$g_2(\cdot)$} & \multicolumn{1}{c|}{$g_2(\cdot)$} & \multicolumn{1}{c|}{$g_3(\cdot)$} \\ \hline
\end{tabular}
\label{tab:confoundingMatrix}
\end{table}

The left plot in Fig.~\ref{fig:fig1} shows the exponential decay over time of the error probability of agent $1$ (similar behavior is observed for the other agents). This exponential decay is in accordance with the bound \eqref{eq:OSLConsistency} from Theorem~\ref{th:consistency}. The example highlights the adaptation-learning trade-off. By examining the curves corresponding to different adaptation parameters, we see that larger values ensure faster adaptation. 
On the other hand, smaller values of the adaptation parameters increase the reaction time, but significantly reduce the steady-state error probability. This behavior is evident from comparing the three curves in Fig.~\ref{fig:fig1}. 

The right plot of Fig.~\ref{fig:fig1} examines the steady-state behavior of the error probability as a function of the adaptation parameters. Specifically, the figure shows that the empirical error probability of agent $1$, evaluated for large $t$, approaches zero for diminishing values of the adaptation parameters, in accordance with \eqref{eq:OSLAsymConsistency} (similar behavior is observed for the other agents).

\begin{figure}[t]
\begin{minipage}{0.45\linewidth}
\centering
\includegraphics[width=0.45\linewidth]{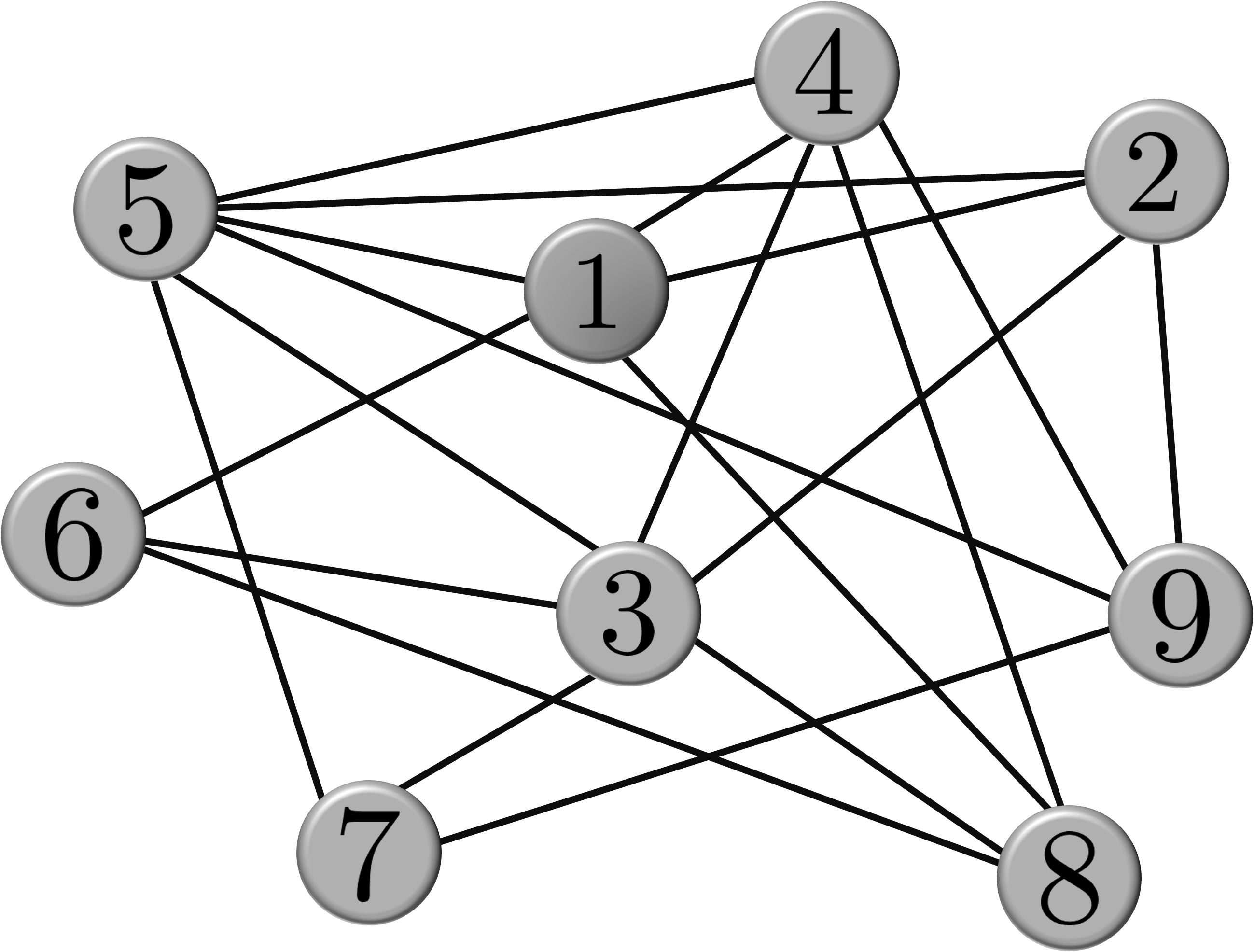}
\end{minipage}
\hfill
\begin{minipage}{0.45\linewidth}
\centering
\includegraphics[width=0.45\linewidth]{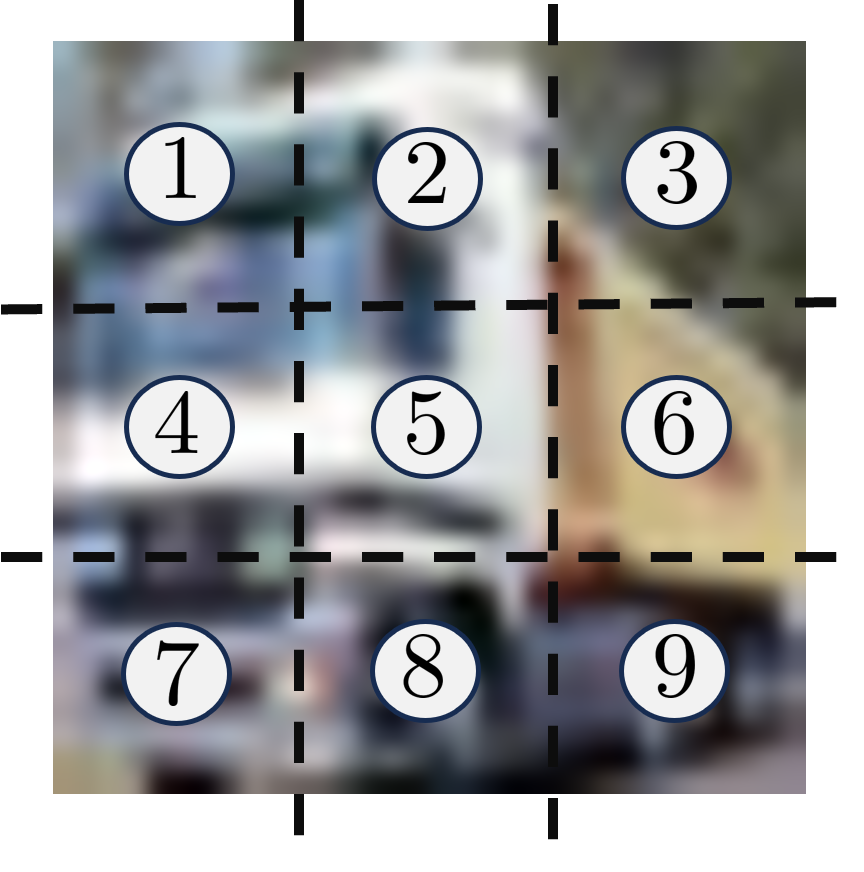}
\end{minipage}
\caption{{\em Left.} Network topology used in the example from Sec.~\ref{sec:exp2}. All agents have a self-loop not shown for ease of illustration. {\em Right.} Example of image patches (from a truck picture) assigned to agents $1,2,\ldots,9$.}
\label{fig:fig2}
\end{figure}

\subsection{Real Data}
\label{sec:exp2}
In this second example we apply the $\asl$ strategy to solve a distributed image classification task, using the popular CIFAR-10 data set~\cite{cifar10}. We reduced the number of classes in the data set to $3$, by extracting from the data set only images of cars, airplanes, and trucks ($1616$ samples per class). From these images we build a purposely biased training set by removing all images containing red cars.

The distributed classification problem is performed over a network of $K=9$ agents, connected according to the topology displayed in the left panel of Fig.~\ref{fig:fig2}. More in detail, given a prediction sample, each agent observes only a patch thereof (right panel of Fig.~\ref{fig:fig2}), and must decide from which class the image was generated. Each agent employs the transformer in~\cite{transformer} as a feature extractor, to map the images into feature vectors of dimension $384\times 1$ that are used for training and/or prediction by the social learning strategy\footnote{The observed patches are resized to match the dimension of the transformer's input and extract the features.}.

\begin{figure}[t]
\centering
\includegraphics[width=0.6\linewidth]{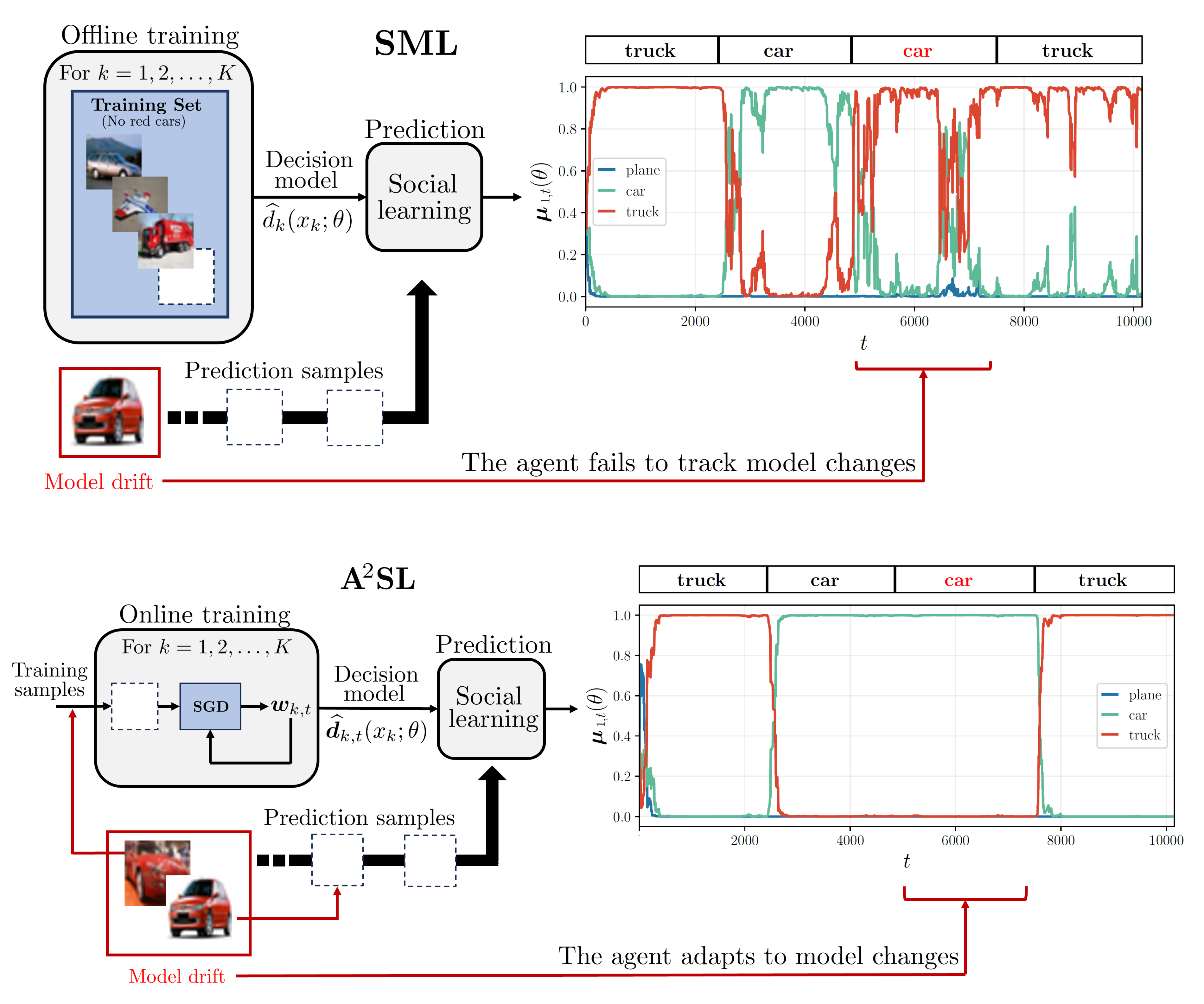}
\caption{Social learning problem over the CIFAR-10 data set~\cite{cifar10}, as illustrated in Sec.~\ref{sec:exp2}. {\em Top}. SML behavior and beliefs evolution of agent $1$ (similar plots for the remaining agents). {\em Bottom}. $\asl$ behavior and beliefs evolution of agent $1$ (similar plots for the remaining agents).}
\label{fig:fig3}
\end{figure}

The aim of the distributed solution is to let agents blend their partial views on the same image, to increase the confidence in their decision. We assume that both the true hypothesis and the likelihood models change over time. We generate the second type of drift by introducing {\em unseen} training and prediction samples of red car images, in order to make the decision model learnt by the agents obsolete. We tackle the distributed classification task under drifting conditions by using the $\asl$ strategy and the social machine learning (SML) strategy~\cite{BordignonVlaski2023,HuBordignon2023}.

We consider the SML strategy since it is the most advanced adaptive social learning strategy that handles unknown models. A scheme of the SML functioning is shown in the top panel of Fig.~\ref{fig:fig3}. 
With this strategy, each agent learns {\em offline} (we use a batch SGD algorithm) the parameter vectors $w_k$ and $u_k$. While SML tracks well drifts in the hypotheses, the agents learn offline the decision statistic and are not able to deal with the model drift. This behavior can be observed in the right plot in the top panel of Fig.~\ref{fig:fig3}. The true hypothesis $\theta_0$ changes two times, at $t\approx2500$ and $t\approx7500$, and the beliefs of agent $1$ evolve and correctly detect $\theta_0$. In contrast, when at time $t \approx 5000$ the likelihood models at all agents drift due to the appearance of red cars in the prediction samples, the decision model learnt offline does not allow agent $1$ to correctly classify the image patches. This behavior is well captured by the belief evolution in the time interval from $t\approx5000$ to $t\approx 6500$. During these iterations of SML, agent $1$ often misclassifies red cars with trucks, since its model does not reflect the observed reality. The model learnt offline, using the initial data set, induces classification errors because the training set contains several images of red trucks, easily confused with red cars.

In the bottom panel of Fig.~\ref{fig:fig3}, we consider the $\asl$ strategy with adaptation parameters $\delta=0.005$, $\eta=0.1$ and $\widetilde{\eta}=0.1$. The regularization parameters for the cross-entropy costs \eqref{eq:genericCrossEntropy} and \eqref{eq:genericCrossEntropyU} are set as $\rho = 0.01$ and $\widetilde{\rho} = 0.01$.  
We see that the beliefs values converge exponentially fast to the truth any time that drifts occur in the true hypothesis {\em or in the models}. Such desirable behavior is enabled by the doubly adaptive update of $\asl$, in particular the SGD updates let the agents track the new decision statistics once the model drift is encountered. Comparing the $\asl$ beliefs evolution of agent $1$ to the SML ones, we see the superiority of the proposed strategy in the online setting. Just after a few iterations required to adapt to the changes, the agents correctly track the true hypothesis even in the presence of the model drift. Indeed, in the time interval from $t\approx5000$ to $t\approx 6500$ agent $1$ does not incur in any misclassifications, irrespective of the model and hypotheses drifts.

\section{Conclusion}
This work considers social learning problems in the fully online setting, where both the true hypothesis and the likelihood models can be subject to drifts over time. Existing algorithms for social learning are not designed to deal with both sources of nonstationarity. Accordingly, we propose the doubly adaptive social learning strategy, in short $\asl$, which is equipped with adaptation capabilities against both drifts. We show that for sufficiently small values of the adaptation parameters, the $\asl$ strategy learns the truth regardless of the possible drift sources. In particular, as time goes on, the agents employing the $\asl$ strategy are characterized by a small probability of error on the order of the adaptation parameters.

\appendices

\section{Proof of Theorem~\ref{th:consistency}}
\label{app:consistencyProof}

\subsection{Useful results on SGD recursions 
\label{app:usefulSGD}
\eqref{eq:trainStep1}--\eqref{eq:trainStep2}}
Consider the error iterates $\bm{w}_{k, t} - w_k^o$ associated with the SGD recursion~\eqref{eq:trainStep1}. In order to obtain the relation in \eqref{eq:asynBasicSGDRecursionW} for the SGD recursion in \eqref{eq:trainStep1}, we first need to introduce the {\em gradient noise}~\cite{Sayed, sayednewbooks} due to using the gradient of the loss function $Q_{k}(w_k;\bm{x}_{k,t}',\bm{\theta}_{k,t}')$ rather than the gradient of the cost function $J_{k}(w_k)$ in~\eqref{eq:genericCrossEntropy}.
\begin{property}[{\bf Gradient noise process}]\label{prop:gradNoise} 
Let the gradient noise be 
\beq 
\bm{s}_{k,t}(\bm{w}_{k,t}) \triangleq \nabla Q_{k}\left(\bm{w}_{k,t};\bm{x}_{k,t}',\bm{\theta}_{k,t}'\right) - \nabla J_{k}(\bm{w}_{k,t}).
\label{eq:whoisgradnoise}
\eeq 
Conditioned on all the iterates $\{ \bm{w}_{k,t-1} \}$ generated up to time $t-1$ by the SGD recursion \eqref{eq:asynBasicSGDRecursionW}, the gradient noise satisfies the properties:
\begin{align}
&\E\big[\bm{s}_{k,t}(\bm{w}_{k,t-1}) \,|\, \{ \bm{w}_{k,t-1} \}
\big] = 0,
\label{eq:gradzeromean}\\ 
&\E\big[\|\bm{s}_{k,t}(\bm{w}_{k,t-1})\|^2 \,|\, \{ \bm{w}_{k,t-1} \}
\big] \leq \sigma^2_{k} ,
\label{eq:gradNoiseUncorrelated}
\end{align}
for some positive constant $\sigma_{k}^2$.
\end{property}
\begin{IEEEproof}
Property \eqref{eq:gradzeromean} holds because $J_k(w_k)$ is defined as the expected value of $Q_k(w_k; \bm{x}_{k,t}', \bm{\theta}_{k,t}')$ taken with respect to $(\bm{x}_{k,t}'$ , $\bm{\theta}_{k,t}')$ -- see \eqref{eq:genericCrossEntropy} -- and these variables are independent of $\bm{w}_{k,t-1}$. Next, we focus on \eqref{eq:gradNoiseUncorrelated}.

Using \eqref{eq:softmaxAppearsAlternative} in \eqref{eq:genericCrossEntropy}, the regularized loss function, for $\bm{x}_{k,t}'=x_k$ and $\bm{\theta}_{k,t}'=\theta$, can be written as
\begin{equation}
    Q_k(w_k;x_k,\theta)=
    -x_k^{\top}w_k(\theta) + \log\sum_{\tau\in\Theta}e^{x_k^{\top}w_k(\tau)}
    +\frac{\rho}{2}\|w_k\|^2.
\label{eq:explicLossQ}
\end{equation}
We recall that the vector $w_k\in\mathbb{R}^{(M_k+1)\times (H-1)}$ is obtained by stacking (columnwise) the $H-1$ vectors $w_k(\theta)\in\mathbb{R}^{M_k+1}$. We denote by $w_k(m,h)$ the $m$-th component of $w_k(\theta_h)$, for $m=1,2,\ldots,M_k+1$ and $h=1,2,\ldots,H-1$. Using this double-index notation, the $(m,h)$ component of the gradient of $Q_k$ in \eqref{eq:explicLossQ} is
\begin{align}
    \frac{\partial Q_k(w_k;x_k,\theta)}{\partial w_k(m,h)}&=
    -x_k(m)\mathbb{I}[\theta=\theta_h] 
    + \frac{x_k(m) e^{x_k^{\top}w_k(\theta_h)}}{\sum\limits_{\tau\in\Theta}e^{x_k^{\top}w_k(\tau)}}
    + \rho \, w_k(m,h),
    \label{eq:morexplicitgradQ}
\end{align}
where $x_k(m)$ denotes the $m$-th component of $x_k$ and $\mathbb{I}$ is the indicator function, which is equal to $1$ if the condition defined by its argument is true and is equal to zero otherwise. 
When evaluating the gradient noise in \eqref{eq:whoisgradnoise}, the last term in \eqref{eq:morexplicitgradQ} disappears. 
The bounded variance condition \eqref{eq:gradNoiseUncorrelated} then follows from the fact that both terms on the RHS of \eqref{eq:morexplicitgradQ} are bounded by $|x_k(m)|$, which has finite second moment in view of Assumption~\ref{ass:trainingBernoulli}.
\end{IEEEproof}
It is known that, in view of Properties~\ref{prop:costFuncProp} and \ref{prop:gradNoise}, and under the characterization of the process $\bm{\alpha}^{\rm{tr}}_{k,t}$ from Assumption~\ref{ass:trainingBernoulli}, for sufficiently small step-size $\eta$, namely, $\eta < (2\,\nu_{k})/\xi^2_{k}$, the iterates $\bm{w}_{k, t}$ obey the following upper bound~\cite{Sayed, sayednewbooks}:
\beq 
\E\| \bm{w}_{k, t} - w_k^{o} \|^2 \leq \varphi_{k}^t\, \E\| \bm{w}_{k,0} - w_k^{o} \|^2 + \eta \,\sigma^2_{k},
\label{eq:sgdMSERecursionW}
\eeq
where
\beq 
\varphi_{k} = 1 - 2\,\eta \, q_k^{\rm{tr}}\nu_{k} + \eta^2\,q_k^{\rm{tr}}\xi_{k}^2
\label{eq:convergenceRate}
\eeq 
is the convergence rate, and $\varphi_{k} \in (0,1]$.
Iterating \eqref{eq:sgdMSERecursionW} and applying the limit superior as $t\rightarrow\infty$ gives the result in~\eqref{eq:MSDclassic}.

Following similar arguments, if $\widetilde{\eta} < (2\,\widetilde{\nu}_{k})/\widetilde{\xi}^2_{k}$ the SGD recursion~\eqref{eq:trainStep2} obeys the bound 
\beq 
\E\| \bm{u}_{k,t} - u_k^{o} \|^2 \leq 
\widetilde{\varphi}_{k}^t \, \E\| \bm{u}_{k,0} - u_k^{o} \|^2 + \widetilde{\eta}\, \widetilde{\sigma}^2_{k}.
\label{eq:sgdMSERecursionPrior}
\eeq 
The convergence rate $\widetilde{\varphi}_{k} \in (0,1]$  
has the same expression as~\eqref{eq:convergenceRate}, with $\widetilde{\eta}$ in place of $\eta$ and the constants $\widetilde{\nu}_{k}$ and $\widetilde{\xi}_{k}$ in place of $\nu_{k}$ and $\xi_{k}$, respectively. 
The steady-state error $\widetilde{\sigma}^2_{k}$ is related to the gradient noise due to using in the SGD recursion~\eqref{eq:trainStep2} the gradient of the loss function $\widetilde{Q}_{k}(u_k;\bm{\theta}_{k,t}')$ rather than the gradient of the cost function $\widetilde{J}_{k}(u_k)$ in~\eqref{eq:genericCrossEntropyU}. 
This second gradient noise satisfies an equivalent version of Property~\ref{prop:gradNoise}, formulated for $\widetilde{J}_{k}(u_k)$ and $\widetilde{Q}_{k}(u_k;\bm{\theta}_{k,t}')$, with the constant $\widetilde{\sigma}^2_{k}$ in place of $\sigma^2_{k}$. Iterating \eqref{eq:sgdMSERecursionPrior}, and applying the limit superior as $t\rightarrow\infty$, Eq.~\eqref{eq:MSDclassic2} is obtained.

\subsection{Useful intermediate result}
The first part of the proof examines the deviation of the log belief ratios from the limit value $\beta_{\rm{net}}(\theta)$, providing an intermediate result that will be used to prove the bound on the instantaneous error probability in~\eqref{eq:OSLConsistency}.

In the derivations we employ the following inequality. Let $\bm{z}_m$ be a random sequence (the same result is also verified for the deterministic case), and let $r \in (0,1)$. Then it holds that: 
\beq 
\E\left[\left( \sum_{m = 0}^{t-1}r^{m} \bm{z}_{m} \right)^2\right] \leq \frac{1 - r^t}{1 - r} \sum_{m=0}^{t-1} r^m\E\,\bm{z}_{m}^2.
\label{eq:gymsenIneq}
\eeq 
We can prove \eqref{eq:gymsenIneq} as follows:
\begin{align}
& \left( \sum_{m = 0}^{t-1}r^{m} \bm{z}_{m} \right)^2 = \left( \frac{1 - r^t}{1 - r} \right)^2 \left( \sum_{m = 0}^{t-1} r^m \frac{1-r}{1-r^t} \bm{z}_m \right)^2 \leq \frac{1 - r^t}{1 - r} \sum_{m = 0}^{t-1} r^m \bm{z}_m^2 ,
\label{eq:middleResultUsefulIneq}
\end{align}
where the inequality follows by Jensen's inequality applied to the square function (note that the  weights $r^m (1-r^t)/(1-r)$, for $m=0,1,\ldots,t,$ are positive and add up to $1$). Taking the expectation of~\eqref{eq:middleResultUsefulIneq} gives~\eqref{eq:gymsenIneq}.

By invoking Markov's inequality, for any $\varepsilon > 0$ we can write:
\beq 
\mathbb{P}\left[|\delta\bm{\beta}_{k,t}(\theta) - \beta_{\rm{net}}(\theta)| \geq \varepsilon\right] \leq \frac{\E\left[\big(\delta\bm{\beta}_{k,t}(\theta) - \beta_{\rm{net}}(\theta)\big)^2\right]}{\varepsilon^2},
\label{eq:markovIneq}
\eeq 
and then focus on characterizing the second order moment on the RHS of \eqref{eq:markovIneq}.
We first write $\beta_{\rm{net}}(\theta)$ as 
\begin{align}
& \beta_{\rm{net}}(\theta) \overset{(\text{I})}{=} \delta\sum_{m=0}^{\infty} (1 - \delta)^m \beta_{\rm{net}}(\theta) \nonumber \\
& \overset{(\text{II})}{=} (1-\delta)^t \beta_{\rm{net}}(\theta) + \sum_{m=0}^{t-1}(1 -\delta)^m \beta_{\rm{net}}(\theta) \nonumber \\
& \overset{(\text{III})}{=}  (1-\delta)^t \beta_{\rm{net}}(\theta) + \sum_{m=0}^{t-1}\sum_{j=1}^K (1 -\delta)^m v_{k} \, q_j^{\rm{pr}}
\left(\E\,\bm{x}_{j}^{\top}\Delta_j^{o}(\theta) - \widetilde{\Delta}_j^{o}(\theta) \right),
\label{eq:betaNetDef}
\end{align}
where steps $(\text{I})$ and $(\text{II})$ follow by the relations $\delta \sum_{m=0}^{\infty} (1 -\delta)^m = 1$ and $\sum_{m=t}^{\infty} (1 -\delta)^m =  (1-\delta)^t/\delta$, respectively, holding for $|\delta| < 1$; step $(\text{III})$ follows by the definition of $\beta_{\rm{net}}(\theta)$ from~\eqref{eq:limitLogBel}.
Using~\eqref{eq:slRecDev} and \eqref{eq:betaNetDef}, along with the definition of the following quantities, 
\begin{align}
& \bm{\Xi}_{j, t}(\theta) \triangleq \bm{\Delta}_{j,t}(\theta) - \Delta_j^{o}(\theta),
\label{eq:wTildeDef}
\\
& \widetilde{\bm{\Xi}}_{j, t}(\theta) \triangleq \widetilde{\bm{\Delta}}_{j,t}(\theta) - \widetilde{\Delta}_j^{o}(\theta),
\label{eq:uTildeDef}
\end{align}
we can write
\begin{align}
\delta\bm{\beta}_{k,t}(\theta) - \beta_{\rm{net}}(\theta) 
& = \delta\bm{\gamma}_t + (1-\delta)^t \beta_{\rm{net}}(\theta) \nonumber \\
& + \sum_{m=0}^{t-1} \sum_{j = 1}^K (1-\delta)^m \Bigg\{[A^{m+1}]_{j k} \bm{\alpha}^{\rm{pr}}_{j,t-m} \Big(\bm{x}_{j,t-m}^{\top}\bm{\Xi}_{j, t-m}(\theta) + \bm{x}_{j,t-m}^{\top}\Delta_j^{o}(\theta) \nonumber \\
&  -  \widetilde{\bm{\Xi}}_{j,t-m}(\theta) - \widetilde{\Delta}_j^{o}(\theta) \Big)  - v_{k} \, q_j^{\rm{pr}}
\left(\E\,\bm{x}_{j}^{\top}\Delta_j^{o}(\theta) - \widetilde{\Delta}_j^{o}(\theta)\right) \Bigg\}.
\label{eq:firstSplitBetaStar} 
\end{align}
By adding and subtracting $(1-\delta)^m[A^{m+1}]_{j k}q_j^{\rm{pr}}\E\bm{x}_j^{\top}\Delta^{o}_j(\theta)$ and $(1-\delta)^m[A^{m+1}]_{j k} q_j^{\rm{pr}}\widetilde{\Delta}^{o}_j(\theta)$ into the summation
we can rearrange the terms in \eqref{eq:firstSplitBetaStar} as
\beq
\delta\bm{\beta}_{k,t}(\theta) - \beta_{\rm{net}}(\theta) = 
\bm{a} + \bm{b} + \bm{c} + d - \bm{e} - \bm{f} - g,
\label{eq:sevenPieces}
\eeq
where
\begingroup
\allowdisplaybreaks
\begin{align}
& \bm{a} = \delta\bm{\gamma}_t + (1-\delta)^t \beta_{\rm{net}}(\theta), 
\label{eq:TDef}
\\
& \bm{b} \!=\! \delta\!\sum_{m=0}^{t-1} \sum_{j = 1}^K  (1-\delta)^m [A^{m+1}]_{j k}\bm{\alpha}^{\rm{pr}}_{j,t-m}\bm{x}_{j,t-m}^{\top}\bm{\Xi}_{j, t-m}(\theta), 
\label{eq:aDef} \\
& \bm{c} = \delta\sum_{m=0}^{t-1} \sum_{j = 1}^K (1-\delta)^m [A^{m+1}]_{j k}  \\ 
& \times \left(\bm{\alpha}^{\rm{pr}}_{j,t-m}\bm{x}_{j,t-m}^{\top} 
- q_j^{\rm{pr}}\E \, \bm{x}_{j}^{\top} \right) \Delta^o_j(\theta),  
\label{eq:bDef}
\\
& d \!=\! \delta\! \sum_{m=0}^{t-1} \sum_{j = 1}^K (1-\delta)^m \left([A^{m+1}]_{j k} - v_j\right) q_j^{\rm{pr}} \E \,\bm{x}_{j}^{\top}\Delta^o_j(\theta), 
\label{eq:cDef}
\\
& \bm{e} = \delta \sum_{m=0}^{t-1}\sum_{j=1}^K (1-\delta)^m [A^{m+1}]_{jk}  \bm{\alpha}^{\rm{pr}}_{j,t-m} \widetilde{\bm{\Xi}}_{j, t-m}(\theta), 
\label{eq:dDef}
\\
& \bm{f} \!=\! \delta\!\sum_{m=0}^{t-1} \sum_{j = 1}^K (1-\delta)^m [A^{m+1}]_{j k} \left(\bm{\alpha}^{\rm{pr}}_{j,t-m}  -  q_j^{\rm{pr}} \right) \widetilde{\Delta}^o_j(\theta), 
\label{eq:eDef}
\\
& g = \delta \sum_{m=0}^{t-1} \sum_{j = 1}^K (1-\delta)^m \left([A^{m+1}]_{j k} - v_j\right) q_j^{\rm{pr}} \widetilde{\Delta}^o_j(\theta).
\label{eq:fDef}
\end{align}
\endgroup
Accordingly, we have, in view of \eqref{eq:sevenPieces},  
\begin{align}
& \E\left[\big(\delta \bm{\beta}_{k,t}(\theta) - \beta_{\rm{net}}(\theta) \big)^2\right] \leq 7\E \,\bm{a}^2 + 7\E\,\bm{b}^2 + 7\E\,\bm{c}^2  + 7d^2  + 7\E\,\bm{e}^2 +7\E\,\bm{f}^2 + 7g^2,
\label{eq:telescopingJensened}
\end{align}
where the inequality is an application of Jensen's inequality with uniform weights equal to $1/7$.

We now examine separately the individual terms in \eqref{eq:telescopingJensened}, starting from $\E \,\bm{a}^2$. Using \eqref{eq:TDef} and the definition of $\bm{\gamma}_t$ from \eqref{eq:slRecDev}, we can write
\beq 
\E\,\bm{a}^2  = \kappa_a(\theta)(1-\delta)^{2t}, 
\eeq 
where
\begin{align}
\kappa_a(\theta) & \triangleq \E\Bigg[\bigg( \sum_{j = 1}^K  \Big[ \delta[A^t]_{j k}\bm{\beta}_{j,0} (\theta) + v_j q_j^{\rm{pr}}\left(\E\,\bm{x}_{j}^{\top}\Delta_j^o(\theta) -    \widetilde{\Delta}_j^o(\theta) \right) \Big] \bigg)^2\Bigg].
\end{align}
Regarding the term $\E\,\bm{b}^2$, using \eqref{eq:gymsenIneq}, \eqref{eq:aDef}, and invoking Assumption~\ref{ass:trainingBernoulli} and the fact that the entries of the matrix $A$ are strictly less than $1$, we have the following upper bound:
\begin{align}
\E\,\bm{b}^2 & \leq \delta\left[1 - (1 - \delta)^t\right]\sum_{m=0}^{t-1}(1-\delta)^m q_j^{\rm{pr}} \E\left[\left(\bm{x}_{j,t-m}^{\top}\bm{\Xi}_{j, t-m}(\theta)\right)^2\right].
\label{eq:firstDecompA}
\end{align}
Applying the Cauchy-Schwarz inequality
we can write
\begin{align}
& \E\left[\left(\bm{x}_{j,t-m}^{\top}\bm{\Xi}_{j, t-m}(\theta)\right)^2\right] \nonumber \\
& \leq \E\| \bm{x}_{j} \|^2 \E\| \bm{\Xi}_{j, t-m}(\theta) \|^2 \nonumber \\
& \leq 2\E\| \bm{x}_{j}\|^2 \big(\E\| \bm{w}_{j, t-m}(\theta_0) - w_j^o(\theta_0) \|^2 + \E\| \bm{w}_{j, t-m}(\theta) - w_j^o(\theta) \|^2 \big),
\label{eq:jensenForMainTerms}
\end{align}
where in the last inequality we applied Jensen's inequality with uniform weights equal to $1/2$ and used the definition in~\eqref{eq:wTildeDef}. Since the squared norm of the $(M_j + 1)\times 1$ vector $\bm{w}_{j, t}(\theta) - w_j^o(\theta)$ is upper bounded by the squared norm of the extended $(M_j + 1)(H-1)\times 1$ vector $\bm{w}_{j, t} - w_j^o$, in view of \eqref{eq:sgdMSERecursionW} we can write
\begin{align}
& \E\left[\left(\bm{x}_{j,t-m}^{\top}\bm{\Xi}_{j, t-m}(\theta)\right)^2\right] \leq 4\E\| \bm{x}_{j} \|^2 \left( \varphi_{j}^{t-m}\E\|\bm{w}_{j, 0} - w_j^o\|^2 + \eta \, \sigma^2_{j} \right).
\label{eq:finalUsefulBoundOnTilde}
\end{align}
Substituting \eqref{eq:finalUsefulBoundOnTilde} in~\eqref{eq:firstDecompA}, we get
\begingroup
\allowdisplaybreaks
\begin{align}
& \E\,\bm{b}^2  \leq \delta\left[1 - (1 - \delta)^t\right] \sum_{m=0}^{t-1} \sum_{j=1}^K (1-\delta)^m \varphi_{j}^{t-m} q_j^{\rm{pr}} 4\E\| \bm{x}_{j} \|^2\E\|\bm{w}_{j, 0} - w_j^o\|^2   \nonumber \\
& + \delta\!\left[1 - (1 - \delta)^t\right]\! \sum_{m=0}^{t-1}  \sum_{j=1}^K (1-\delta)^m q_j^{\rm{pr}} 4\E\| \bm{x}_{j} \|^2 \eta \, \sigma^2_{j} \nonumber \\
& \overset{(\text{I})}{\leq} \delta\varphi^t\left[1 - (1 - \delta)^t\right] \sum_{m=0}^{t-1} \sum_{j=1}^K \underbrace{\left(\frac{1-\delta}{\varphi_{j}}\right)^m}_{\triangleq \chi_{j}^m} q_j^{\rm{pr}} 4\E\| \bm{x}_{j} \|^2\E\|\bm{w}_{j, 0} - w_j^o\|^2 \nonumber \\
& + \eta \left[1 - (1 - \delta)^t\right]^2\sum_{j=1}^K q_j^{\rm{pr}}4\E\| \bm{x}_{j} \|^2 \sigma^2_{j} \nonumber \\
& \overset{(\text{II})}{\leq} \delta\varphi^t\left[1 - (1 - \delta)^t\right] \sum_{j=1}^{K}\frac{1 - \chi_{j}^t}{1-\chi_{j}} \sum_{j=1}^K q_j^{\rm{pr}} 4\E\| \bm{x}_{j} \|^2\E\|\bm{w}_{j, 0} - w_j^o\|^2 \nonumber \\
& + \eta \left[1 - (1 - \delta)^t\right]^2\sum_{j=1}^K q_j^{\rm{pr}}4\E\| \bm{x}_{j} \|^2 \sigma^2_{j},
\end{align}
\endgroup
where step $(\text{I})$ follows by~\eqref{eq:slowestConvRates}, and step $(\text{II})$ follows by computing the geometric sum with coefficient $\chi_{j}$, which is convergent since $\chi_{j} < 1$ under condition \eqref{eq:stabConditionEta}.
So we can conclude that 
\begingroup
\allowdisplaybreaks
\begin{align}
& \E\,\bm{b}^2  \leq \delta\varphi^t \left[1 - (1 - \delta)^t\right] \underbrace{\sum_{j=1}^K 4 \frac{1 - \chi_{j}^t}{1-\chi_{j}} q_j^{\rm{pr}} \E\| \bm{x}_{j} \|^2\E\|\bm{w}_{j, 0} - w_j^o\|^2}_{\triangleq \kappa^{(1)}_{b}(\theta)} \nonumber \\
& + \eta \left[1 - (1 - \delta)^t\right]^2\underbrace{\sum_{j=1}^K q_j^{\rm{pr}}4\E\| \bm{x}_{j} \|^2 \sigma^2_{j}}_{\triangleq \kappa^{(2)}_{b}(\theta)} \nonumber \\
& \leq \varphi^t \delta\, \kappa^{(1)}_{b}(\theta) + \eta \, \kappa^{(2)}_{b}(\theta).
\label{eq:gradTermDecomp}
\end{align}
\endgroup
Using \eqref{eq:bDef}, the term $\E\,\bm{c}^2$ can be upper bounded as:
\begingroup
\allowdisplaybreaks
\begin{align}
& \E\,\bm{c}^2  \overset{(\text{I})}{\leq} \delta^2 \E\Bigg[\bigg( \sum_{m=0}^{t-1} \sum_{j = 1}^K (1-\delta)^m \left(\bm{\alpha}^{\rm{pr}}_{j,t-m}\bm{x}_{j,t-m}^{\top} 
- q_j^{\rm{pr}}\E \,\bm{x}_{j,t-m}^{\top}\right) \Delta^o_j(\theta) \bigg)^2\Bigg] \nonumber \\
& \overset{(\text{II})}{=} \delta^2 \sum_{m=0}^{t-1} \sum_{j = 1}^K (1-\delta)^{2m} \E\left[\left( \left(\bm{\alpha}^{\rm{pr}}_{j,t-m}\bm{x}_{j,t-m}^{\top} 
- q_j^{\rm{pr}}\E\,\bm{x}_{j,t-m}^{\top}\right) \Delta^o_j(\theta)  \right)^2\right] \nonumber \\
& \overset{(\text{III})}{\leq} \delta^2 \sum_{m=0}^{t-1} \sum_{j = 1}^K (1-\delta)^{2m} \E\left\|\bm{\alpha}^{\rm{pr}}_{j,t-m}\bm{x}_{j,t-m}^{\top} 
- q_j^{\rm{pr}}\E\,\bm{x}_{j,t-m}^{\top}\right\|^2\left\|\Delta^o_j(\theta)\right\|^2 \nonumber \\
& = \delta\!\left[\frac{1- (1-\delta)^{2t}}{2-\delta} \sum_{j=1}^K \sum_{\ell=1}^{M_j} \textnormal{Var}\left(\bm{\alpha}^{\rm{pr}}_{j}[\bm{x}_{j}]_{\ell}\right) \|\Delta^o_j(\theta)\|^2 \right] \nonumber \\
& \leq \delta\, \kappa_c(\theta),
\label{eq:cTermDecomp}
\end{align}
\endgroup
where
\beq 
\kappa_c(\theta) \triangleq \frac{1}{2-\delta}\sum_{j=1}^K \sum_{\ell=1}^{M_j} \textnormal{Var}\left(\bm{\alpha}^{\rm{pr}}_{j}[\bm{x}_{j}]_{\ell}\right) \|\Delta^o_j(\theta)\|^2 ,
\eeq 
and where step $(\text{I})$ follows by the fact that the entries of the combination matrix are strictly less than $1$, step $(\text{II})$ stems from the zero mean property of the individual terms of the summation, and step $(\text{III})$ is an application of the Cauchy-Schwarz inequality. 
For the term $d^2$, using \eqref{eq:cDef} we have: 
\begingroup
\allowdisplaybreaks
\begin{align}
& d^2 \overset{(\text{I})}{\leq} \delta^2 \E\left[\left(  \sum_{m=0}^{t-1} \sum_{j = 1}^K \kappa [(1-\delta)\zeta]^m  q_j^{\rm{pr}} \E \,\bm{x}_{j}^{\top}\Delta^o_j(\theta) \right)^2\right] \nonumber \\
& \overset{(\text{II})}{\leq} \delta\frac{\left[1 - [(1 - \delta)\zeta]^t\right]}{1 - (1-\delta)\zeta} \kappa^2 \sum_{m=0}^{t-1}  \sum_{j=1}^K [(1-\delta)\zeta]^m q_j^{\rm{pr}} \left( \E \, \bm{x}_{j}^{\top}\Delta^o_j(\theta) \right)^2 \nonumber \\
& \overset{(\text{III})}{\leq} \delta\!\left[\!\left(\frac{\left[1 - [(1 - \delta)\zeta]^t\right]}{1 - (1-\delta)\zeta} \right)^2 \!\!\! \kappa^2 \! \sum_{j=1}^K \! q_j^{\rm{pr}}\| \E \, \bm{x}_{j}\|^2 \| \Delta^o_j(\theta) \|^2\!\right] \nonumber \\
& \leq \delta\, \kappa_d(\theta),
\label{eq:dTermDecomp}
\end{align}
\endgroup
where
\beq 
\kappa_d(\theta) \triangleq \frac{\kappa^2}{(1 - (1-\delta)\zeta)^2}\sum_{j=1}^K  q_j^{\rm{pr}}\| \E \, \bm{x}_j^{\top}\|^2 \| \Delta^o_j(\theta) \|^2,
\eeq 
and where step $(\text{I})$ follows by \eqref{eq:matrixPowBound}, step $(\text{II})$ uses \eqref{eq:gymsenIneq}, and step $(\text{III})$ is an application of Cauchy-Schwarz inequality.
We now upper bound the terms $\E\,\bm{e}^2$, $\E\,\bm{f}^2$, and $g^2$ using their definitions in \eqref{eq:dDef}-\eqref{eq:fDef}. Applying the same steps used in~\eqref{eq:firstDecompA}-\eqref{eq:gradTermDecomp} we can write
\beq
\E\,\bm{e}^2 \leq \widetilde{\varphi}^t \delta\, \kappa^{(1)}_e(\theta) + \widetilde{\eta} \, \kappa^{(2)}_e(\theta)
\label{eq:firstPriorTermBound}
\eeq
under condition \eqref{eq:stabConditionEta}, and where 
\begin{align}
& \kappa^{(1)}_e(\theta) \triangleq \sum_{j=1}^K 4 \frac{1 - \widetilde{\chi}_{j}^t}{1-\widetilde{\chi}_{j}} q_j^{\rm{pr}}\E\|\bm{u}_{j,0} - u_j^o\|^2, \\
& \widetilde{\chi}_{j} \triangleq (1-\delta)/\widetilde{\varphi}_{j}, \\
& \kappa^{(2)}_e(\theta) \triangleq \sum_{j=1}^K 4q_j^{\rm{pr}}\widetilde{\sigma}^2_{j}.
\end{align} 
Likewise, applying the same steps that lead to the bounds \eqref{eq:cTermDecomp} and \eqref{eq:dTermDecomp}, respectively, we have that 
$\E\,\bm{f}^2 \leq \delta\, \kappa_f(\theta)$,
where 
\beq 
\kappa_f(\theta) \triangleq \frac{1}{2-\delta}\sum_{j=1}^K \textnormal{Var}\left(\bm{\alpha}_j^{\rm{pr}}\right) \|\widetilde{\Delta}^o_j(\theta) \|^2,
\eeq 
and $g^2 \leq \delta\, \kappa_g(\theta)$,
where
\beq 
\kappa_g(\theta) \triangleq \frac{\kappa^2}{(1 - (1-\delta)\zeta)^2}\sum_{j=1}^K  q_j^{\rm{pr}} \| \widetilde{\Delta}^o_j(\theta)  \|^2.
\eeq 
Plugging into \eqref{eq:telescopingJensened} the bounds obtained for the quantities $\E\,\bm{a}^2$ to $g^2$, we have that 
\begin{align}
& \E\left[\left(\delta \bm{\beta}_{k,t}(\theta) - \beta_{\rm{net}}(\theta)\right)^2\right] \nonumber \\
& \leq 
\kappa^{(1)}_b(\theta)\,\delta\,\varphi^t   
+ 
\kappa^{(1)}_e(\theta) \,\delta\,\widetilde\varphi^t + 
\kappa_a(\theta)(1-\delta)^{2t}  \nonumber \\
& +  \big(\kappa_c(\theta)+\kappa_d(\theta)+\kappa_f(\theta)+\kappa_g(\theta)\big) \delta\nonumber\\
&+ 
\kappa^{(2)}_b(\theta) \eta 
+ 
\kappa^{(2)}_e(\theta) \widetilde{\eta}. 
\label{eq:secondOrder4Markov}
\end{align} 

\subsection{Proof of bound~\eqref{eq:OSLConsistency}}
The instantaneous error probability can be upper bounded by the union bound as
\begin{align}
p_{k,t} & = \mathbb{P}\left[ \exists \theta \neq \theta_0 : \bm{\beta}_{k,t}(\theta) \leq 0 \right] \nonumber \\
& \leq \sum_{\theta \neq \theta_0} \mathbb{P}\left[ \bm{\beta}_{k,t}(\theta) \leq 0 \right] =  \sum_{\theta \neq \theta_0} \mathbb{P}\left[ \delta\bm{\beta}_{k,t}(\theta) \leq 0 \right] \nonumber \\
& = \sum_{\theta \neq \theta_0} \mathbb{P}\left[ \delta\bm{\beta}_{k,t}(\theta) - \beta_{\rm{net}}(\theta) \leq - \beta_{\rm{net}}(\theta) \right].
\label{eq:unionBoundApp}
\end{align}
Then we consider the following probability:
\begin{align}
& \mathbb{P}\left[ |\delta\bm{\beta}_{k,t}(\theta) - \beta_{\rm{net}}(\theta)| \geq \beta_{\rm{net}}(\theta) \right] \nonumber \\
& = \mathbb{P}\left[ \delta\bm{\beta}_{k,t}(\theta) - \beta_{\rm{net}}(\theta) \geq \beta_{\rm{net}}(\theta) \right] + \mathbb{P}\left[ \delta\bm{\beta}_{k,t}(\theta) - \beta_{\rm{net}}(\theta) \leq -\beta_{\rm{net}}(\theta) \right],
\end{align}
which, in view of \eqref{eq:unionBoundApp}, implies that 
\begin{align}
p_{k,t} \leq \sum_{\theta \neq \theta_0} \mathbb{P}\left[ |\delta\bm{\beta}_{k,t}(\theta) - \beta_{\rm{net}}(\theta)| \geq \beta_{\rm{net}}(\theta) \right].
\label{eq:unionBoundSubs}
\end{align}
The $H-1$ probabilities in the summation in \eqref{eq:unionBoundSubs} can be upper bounded using \eqref{eq:markovIneq} and \eqref{eq:secondOrder4Markov} by choosing $\varepsilon = \beta_{\rm{net}}(\theta)$. Therefore, we get the claim in \eqref{eq:OSLConsistency} by joining the $H-1$ bounds in \eqref{eq:unionBoundSubs} and introducing the quantities
\begin{align}
& c_1 \triangleq \max_{\theta \neq \theta_0} \frac{\kappa^{(1)}_b(\theta)}{\beta^2_{\rm{net}}(\theta)},\;  
c_2 \triangleq \max_{\theta \neq \theta_0} \frac{\kappa^{(1)}_e(\theta)}{\beta^2_{\rm{net}}(\theta)},\; 
c_3 \triangleq \max_{\theta \neq \theta_0} \frac{\kappa_a(\theta)}{\beta^2_{\rm{net}}(\theta)},
\nonumber\\
& c_4 \triangleq \max_{\theta \neq \theta_0} \frac{\kappa_c(\theta)+\kappa_d(\theta)+\kappa_f(\theta)+\kappa_g(\theta)}{\beta^2_{\rm{net}}(\theta)},\nonumber\\
& c_5 \triangleq \max_{\theta \neq \theta_0} \frac{\kappa^{(2)}_b(\theta)}{\beta^2_{\rm{net}}(\theta)}, \qquad 
c_6 \triangleq \max_{\theta \neq \theta_0} \frac{\kappa^{(2)}_e(\theta)}{\beta^2_{\rm{net}}(\theta)},
\end{align}
which completes the proof. 



\begin{thebibliography}{10}
\providecommand{\url}[1]{#1}
\csname url@samestyle\endcsname
\providecommand{\newblock}{\relax}
\providecommand{\bibinfo}[2]{#2}
\providecommand{\BIBentrySTDinterwordspacing}{\spaceskip=0pt\relax}
\providecommand{\BIBentryALTinterwordstretchfactor}{4}
\providecommand{\BIBentryALTinterwordspacing}{\spaceskip=\fontdimen2\font plus
\BIBentryALTinterwordstretchfactor\fontdimen3\font minus
  \fontdimen4\font\relax}
\providecommand{\BIBforeignlanguage}[2]{{%
\expandafter\ifx\csname l@#1\endcsname\relax
\typeout{** WARNING: IEEEtran.bst: No hyphenation pattern has been}%
\typeout{** loaded for the language `#1'. Using the pattern for}%
\typeout{** the default language instead.}%
\else
\language=\csname l@#1\endcsname
\fi
#2}}
\providecommand{\BIBdecl}{\relax}
\BIBdecl

\bibitem{A2SLICASSP2024}
M.~Carpentiero, V.~Bordignon, V.~Matta and A.~H. Sayed, ``Social learning with adaptive models,'' in \emph{Proc. IEEE International Conference on Acoustics, Speech, and Signal Processing (ICASSP)}, Seoul, South Korea, Apr. 2024, pp. 9351--9355.

%%%%%%%%%%%%%%%%%%%%%%%%%%%%%%%%%%% NEW MAG and BOOK

\bibitem{SLBook}
 V.~Matta, V.~Bordignon, and A.~H. Sayed, \emph{Social Learning: Opinion Formation and Decision-Making Over Graphs}. Now Publishers, 2025.

\bibitem{SLMagazine2024}
V.~Bordignon, V.~Matta and A.~H. Sayed, ``Socially Intelligent Networks: A framework for decision making over graphs,'' \emph{IEEE Signal Process. Mag.}, vol.~41, no.~4, pp. 20--39, July 2024.

%%%%%%%%%%%%%%%%%%%%%%%%%%%%%% GENERAL
\bibitem{ChamleyBook}
C.~Chamley, \emph{Rational Herds: Economic Models of Social Learning}.\hskip 1em plus 0.5em minus 0.4em\relax Cambridge {U}niversity {P}ress, 2004.

\bibitem{acemoglu2011opinion}
D.~Acemoglu and A.~Ozdaglar, ``Opinion dynamics and learning in social networks,'' \emph{Dynamic Games and Applications}, vol.~1, no.~1, pp. 3--49,
  2011.

\bibitem{acemoglu2011bayesian}
D.~Acemoglu, M.~A. Dahleh, I.~Lobel, and A.~Ozdaglar, ``Bayesian learning in social networks,'' \emph{The Review of Economic Studies}, vol.~78, no.~4, pp. 1201--1236, 2011.

\bibitem{chamley2013models}
C.~Chamley, A.~Scaglione, and L.~Li, ``Models for the diffusion of beliefs in social networks: An overview,'' \emph{IEEE Signal Process. Mag.}, vol.~30, no.~3, pp. 16--29, 2013.

\bibitem{krishnamurthy2013social}
V.~Krishnamurthy and H.~V. Poor, ``Social learning and {B}ayesian games in multiagent signal processing: How do local and global decision makers interact?,'' \emph{IEEE Signal Process. Mag.}, vol.~30, no.~3, pp. 43--57, 2013.

\bibitem{mossel2017opinion}
E.~Mossel and O.~Tamuz, ``Opinion exchange dynamics,'' \emph{Probability Surveys}, vol.~14, pp. 155--204, 2017.

%%%%%%%%%%%%%%%%%%%%%%%%%%%%%%%%%%%% SL
\bibitem{zhao2012learning}
X.~Zhao and A.~H. Sayed, ``Learning over social networks via diffusion adaptation,'' in \emph{Proc. Asilomar Conference on Signals, Systems, and Computers}, Pacific Grove, CA, USA, Nov. 2012, pp. 709--713.

\bibitem{jadbabaie2012non}
A.~Jadbabaie, P.~Molavi, A.~Sandroni, and A.~Tahbaz-Salehi, ``Non-{B}ayesian social learning,'' \emph{Games and Economic Behavior}, vol.~76, no.~1, pp. 210--225, 2012.

\bibitem{ShahrampourTAC2016}
S.~Shahrampour, A.~Rakhlin, and A.~Jadbabaie, ``Distributed detection: Finite-Time analysis and impact of network topology,'' \emph{IEEE Trans. Autom. Control}, vol.~61, no.~11, pp. 3256--3268, 2016.

\bibitem{nedic2017}
A.~Nedi{\'c}, A.~Olshevsky, and C.~A. Uribe, ``Fast convergence rates for distributed non-{B}ayesian learning,'' \emph{IEEE Trans. Autom. Control}, vol.~62, no.~11, pp. 5538--5553, 2017.

\bibitem{molavi2018theory}
P.~Molavi, A.~Tahbaz-Salehi, and A.~Jadbabaie, ``A theory of non-{B}ayesian social learning,'' \emph{Econometrica}, vol.~86, no.~2, pp. 445--490, 2018.

\bibitem{lalitha2018}
A.~Lalitha, T.~Javidi, and A.~D. Sarwate, ``Social learning and distributed hypothesis testing,'' \emph{IEEE Trans. Inf. Theory}, vol.~64, no.~9, pp. 6161--6179, 2018.

%\bibitem{matta2020}
%V.~Matta, V.~Bordignon, A.~Santos, and A.~H. Sayed, ``Interplay between topology and social learning over weak graphs,'' \emph{IEEE Open J. Signal Process.}, vol.~1, pp. 99--119, 2020.

%\bibitem{salami2017social}
%H.~Salami, B.~Ying, and A.~H. Sayed, ``Social learning over weakly connected graphs,'' \emph{IEEE Trans. Signal Inf. Process. Netw.}, vol.~3, no.~2, pp. 222--238, 2017.

%%%%%%%%%%%%%%%%%%%%%%%%%%%%%%%%% ASL

%\bibitem{ASLEusipco2020}
%V.~Bordignon, V.~Matta, and A.~H. Sayed,
%``Adaptation in Online Social Learning,'' in \emph{Proc. European Signal Processing Conference (EUSIPCO)}, Amsterdam, 
%Netherlands, Jan. 2021, pp. 2170--2174.

\bibitem{bordignon2021adaptive}
V.~Bordignon, V.~Matta, and A.~H. Sayed, ``Adaptive social learning,'' \emph{IEEE Trans. Inf. Theory}, vol.~67, no.~9, pp. 6053--6081, 2021.

%\bibitem{PingBordignon2023}
%P.~Hu, V.~Bordignon, S.~Vlaski, and A.~H. Sayed, ``Optimal aggregation strategies for social learning over graphs,'' \emph{IEEE Trans. Inf. Theory},  vol. 69, no. 9, pp. 6048--6070, 2023. 

%%%%%%%%%%%%%%%%%%%%%%%%%%%%%%%%% SML
%\bibitem{BordignonVlaski2021}
%V.~Bordignon, S.~Vlaski, V.~Matta and A.~H. Sayed, ``Network classifiers based on social learning,'' in \emph{Proc. IEEE International Conference on Acoustics, Speech, and Signal Processing (ICASSP)}, Toronto, Canada, June 2021, pp. 5185–-5189.

\bibitem{BordignonVlaski2023}
V.~Bordignon, S.~Vlaski, V.~Matta and A.~H. Sayed, ``Learning from heterogeneous data based on social interactions over graphs,'' \emph{IEEE Trans. on Inf. Theory}, vol. 69, no. 5, pp. 3347--3371, 2023.

\bibitem{HuBordignon2023}
P.~Hu, V.~Bordignon, M.~Kayaalp, and A.~H. Sayed, ``Non-asymptotic performance of social machine learning under limited data,'' \emph{Signal Processing}, vol. 230, 2025.

%%%%%%%%%%%%%%%%%%%%%%%%%%%%%%% UNCERTAIN MODELS
\bibitem{HareUribeTSP2020}
J.~Z. Hare, C.~A. Uribe, L.~Kaplan, and A.~Jadbabaie, ``Non-Bayesian
social learning with uncertain models,'', \emph{IEEE Trans. on Signal
Process.}, vol.~68, pp. 4178--4193, 2020.

\bibitem{cifar10}
A.~Krizhevsky, V.~Nair and G.~Hinton, ``Learning Multiple Layers of Features from Tiny Images,'' 2009 [Online]. Available: https://www.cs.toronto.edu/~kriz/cifar.html.

%%%%%%%%%%%%%%%%%%%%%%%%%%%%% SAYED BOOKS

\bibitem{Sayed}
A.~H. Sayed, ``Adaptation, Learning, and Optimization over Networks,'' \emph{Found. Trends Mach. Learn.}, vol.~7, no. 4-5, pp. 311--801, 2014.

\bibitem{sayednewbooks}
A.~H. Sayed, \emph{Inference and Learning from Data}. Cambridge University Press, 2022.

\bibitem{bib:matrix}
R.~A. Horn and C.~R. Johnson, \emph{Matrix Analysis}. Cambridge {U}niversity {P}ress, 2012.

\bibitem{Meyer}
C.~D. Meyer, \emph{Matrix Analysis and Applied Linear Algebra}. SIAM, 2000.

%%%%%%%%%%%%%%%%%%%%%%%%%%%%%%%%%%% GEN vs DISC
\bibitem{NgJordanNIPS2001}
A.~Ng, and M.~I. Jordan, ``On discriminative vs. generative classifiers: A comparison of logistic regression and na\"ive Bayes,'' in \emph{Proc. Advances on Neural Information Processing Systems (NIPS)}, Vancouver, Canada, Dec. 2001.

\bibitem{VapnikStatLearn}
V.~Vapnik, \emph{The Nature of Statistical Learning Theory}. Springer, 1998.


\bibitem{PatternRecognition}
C.~M. Bishop and N.~M. Nasrabadi, \emph{Pattern Recognition and Machine Learning}. Springer, 2006.

%\bibitem{cover1999elements}
%T.~M. Cover and J.~A. Thomas, \emph{Elements of Information Theory}.\hskip 1em plus 0.5em minus 0.4em\relax Wiley, 1991.

\bibitem{transformer}
A.~Dosovitskiy et al., ``An image is worth 16x16 words: Transformers for image recognition at scale,'' available online as 	arXiv:2010.11929 [cs.CV].

%\bibitem{ZhaoSayedSP2018}
%A.~H. Sayed and X.~Zhao, ``Asynchronous adaptive networks,'' in \emph{Cooperative and Graph Signal Processing}, P.~Djuric and C.~Richard, Eds.\hskip 1em plus 0.5em minus 0.4em\relax Elsevier, 2018, pp. 69--106.



\end{thebibliography}
\end{document}